\documentclass{article}

\PassOptionsToPackage{numbers, compress}{natbib}
\usepackage[final]{neurips_2024}

\usepackage[utf8]{inputenc} 
\usepackage[T1]{fontenc}    
\usepackage{hyperref}       
\usepackage{url}            
\usepackage{booktabs}       
\usepackage{amsfonts}       
\usepackage{nicefrac}       
\usepackage{microtype}      
\usepackage{xcolor}         
\usepackage{multicol}
\usepackage{enumitem}
\usepackage{subfigure}
\usepackage{pdfpages}

\usepackage{graphicx}
\usepackage{caption}
\usepackage{subcaption}
\usepackage{microtype}
\usepackage{comment}
\usepackage{amsmath}
\usepackage{cleveref}
\definecolor{darkblue}{rgb}{0, 0, 0.5}
\hypersetup{colorlinks=true, citecolor=darkblue, linkcolor=darkblue, urlcolor=darkblue}

\newcommand{\citeposs}[1]{\citeauthor{#1}'s \cite{#1}}

\title{LLM Circuit Analyses Are Consistent Across Training and Scale}

\author{Curt Tigges\\
EleutherAI\\
\texttt{curt@eleuther.ai} \\
\And
Michael Hanna\\
ILLC, University of Amsterdam \\
\texttt{m.w.hanna@uva.nl}\\
\And
Qinan Yu\\
Brown University \\
\texttt{qinan\_yu@brown.edu}
\And
Stella Biderman\\
EleutherAI\\
\texttt{stella@eleuther.ai}
}

\begin{document}

\maketitle

\begin{abstract}
Most currently deployed LLMs undergo continuous training or additional finetuning. By contrast, most research into LLMs' internal mechanisms focuses on models at one snapshot in time (the end of pre-training), raising the question of whether their results generalize to real-world settings. Existing studies of mechanisms over time focus on encoder-only or toy models, which differ significantly from most deployed models. In this study, we track how model mechanisms, operationalized as circuits, emerge and evolve across 300 billion tokens of training in decoder-only LLMs, in models ranging from 70 million to 2.8 billion parameters. We find that task abilities and the functional components that support them emerge consistently at similar token counts across scale. Moreover, although such components may be implemented by different attention heads over time, the overarching algorithm that they implement remains. Surprisingly, both these algorithms and the types of components involved therein tend to replicate across model scale. Finally, we find that circuit size correlates with model size and can fluctuate considerably over time even when the same algorithm is implemented. These results suggest that circuit analyses conducted on small models at the end of pre-training can provide insights that still apply after additional training and over model scale.

\end{abstract}

\section{Introduction} \label{sec:intro}
As LLMs' capabilities have grown, so has interest in characterizing their mechanisms. Recent work in mechanistic interpretability often seeks to do so via circuits: computational subgraphs that explain task-solving mechanisms \citep{wang2023interpretability,hanna2023how,merullo2024circuit, lieberum2023does}. Circuits can be found and verified using a variety of methods, \citep{conmy2023towards, syed2023attribution, hanna2024circuit,kramar2024atp,ferrando2024information} with the aim of reverse-engineering models' task-solving algorithms.

Though much circuits research is motivated by LLMs' capabilities, the setting in which such research is performed often differs from that of currently deployed models. Crucially, while most LLM circuits work \citep{wang2023interpretability,hanna2023how,merullo2024circuit, lieberum2023does, tigges2023linear} studies models at the end of pre-training, currently deployed models often undergo continuous training \citep{openai2024gpt4, anthropic2024claude, geminiteam2024gemini} or are fine-tuned for specific tasks \citep{chung2022scaling, hu2021lora}. More generally, models are trained with varying amounts of data and circuits can be sampled at any checkpoint; to what extent do those circuits hold as training continues? Other subfields of interpretability have studied model development during training \citep{hu2023latent, chang2023characterizing, warstadt-etal-2020-learning, choshen-etal-2022-grammar, chang-bergen-2022-word}, but similar work on LLM mechanisms is scarce. Existing mechanistic work over training has studied syntactic attention structures and induction heads \citep{olsson2022context,chen2024sudden, singh2024needs}, but has focused on small encoder or toy models. \citet{prakash2024finetuning} examines circuits in 7-billion-parameter models post-finetuning, but the evolution of circuits during pre-training remains unexplored. This raises questions about whether circuit analyses will generalize if the model in question is further trained on a wide distribution of data. 


We address this issue by exploring when and how circuits and their components emerge during training, and their consistency across training and different model scales.\footnote{Though we do not study fine-tuning or other post-training techniques, observing the types of changes that can occur over the course of pre-training may shed light on the changes that can occur in circuits more generally.} We study circuits in models from the Pythia suite \citep{biderman2023pythia} across 300 billion tokens, at scales from 70 million to 2.8 billion parameters.
We supplement this with additional data from models ranging up to 12 billion parameters. Our results suggest remarkable consistency in circuits and their attributes across scale and training. We summarize our contributions as follows:

 \textbf{Performance acquisition and functional component emergence are similar across scale:} Task ability acquisition rates tend to reach a maximum at similar token counts across different model sizes. Functional components like name mover heads, copy suppression heads, and successor heads also emerge consistently at similar points across scales, paralleling previous findings that induction heads emerge at roughly 2B-5B tokens across models of all scales \citep{olsson2022context}.

 \textbf{Circuit algorithms often remain stable despite component-level fluctuations:} Analysis of the IOI circuit across training and scale reveals that even when individual components change, the overall algorithm remains consistent, indicating a certain level of algorithmic stability. In addition to stability across time, we find that the algorithm also tends to be similar for dramatically different model scales, suggesting that some currently-identified circuits may generalize.

 \textbf{Graph-level circuit attributes vary across training but correlate with model size:} Once primary functionalities emerge, the circuit subgraph constituents tend to stabilize (with circuits in larger models showing higher stability), but we also observe exceptions: constituents can shift significantly even late in training. Circuits in larger models require more components, with circuit sizes positively correlating with model scale.

\textbf{Taken as a whole, our results suggest that circuit analysis generalizes well} over both training and scale even in the face of component and circuit size changes, and that circuits studied at the end of training in smaller models can indeed be informative for larger models as well as for models with longer training runs. We hope to see this validated for other circuits, especially more complex ones and those based on SAE latents, confirming our initial findings.

\section{Methods} \label{section:methods}

\subsection{Circuits}\label{sec:circuits}
A \textbf{circuit} \citep{olah2020zoom,elhage2021mathematical,wang2023interpretability} is the minimal computational subgraph of a model that is faithful to its behavior on a given task. At a high level, this means that circuits describe the components of a model that the model uses to perform the task; in this paper, the components that we study, and thus the nodes in our circuit, are the model's attention heads and multi-layer perceptrons (MLPs). A task, within the circuits framework, is defined by inputs, expected outputs, and a (continuous) metric that measures model performance on the task. For example, in the indirect object identification (IOI, \citep{wang2023interpretability}) task, the LM receives inputs like ``When John and Mary went to the store, John gave a drink to'', and is expected to output \textit{Mary}, rather than \textit{John}. We can measure the extent to which the LM fulfills our expectations by measuring the difference in logits assigned to \textit{Mary} and \textit{John}.

Circuits are useful objects of study because we can verify that they are \textit{faithful} to LM behavior on the given task. We say that a circuit is faithful if we can corrupt all nodes and edges outside the circuit without changing model behavior on the task. Concretely, we test faithfulness by running the model on normal input, while replacing the activations corresponding to edges outside our circuit, with activations from a corrupted input, which elicits very different model behavior. In the above case, our corrupted input could instead be ``When John and Mary went to the store, Mary gave a drink to'', eliciting \textit{John} over \textit{Mary}. If the circuit still predicts \textit{Mary}, rather than \textit{John}, it is faithful. As circuits are often small, including less than 5\% of model edges, this faithfulness test corrupts most of the model, thus guaranteeing that circuits capture a small set of task-relevant model mechanisms. For more details on the circuits framework, see prior work and surveys \citep{conmy2023towards,hanna2024circuit,ferrando2024primer}.

Circuits have a number of advantages over other interpretability frameworks. As computational subgraphs of the model that flow from its inputs to its outputs, they provide complete explanations for a model's mechanisms. Moreover, their faithfulness, verified using a causal test, makes them more reliable explanations. This stands in contrast to probing \citep{belinkov-2022-probing}, which only offers layer-representation-level explanations, and can be unfaithful, capturing features unused by the model \citep{Elazar2020AmnesicPB}. Similarly, input attributions \citep{pmlr-v70-shrikumar17a,Sundararajan2017AxiomaticAF} only address which input tokens are used, and may be unreliable \citep{Adebayo2018SanityCF, bilodeau2024impossibility}.

\subsection{Circuit Finding} \label{sec:circuit-finding}
In order to find faithful circuits at scale over many checkpoints, we use efficient, attribution-based circuit finding methods. Such methods score the importance of all edges in a model's graph in a fixed number of forward and backward passes, independent of model size; though other patching-based circuit-finding methods \citep{conmy2023towards} are more accurate, they are too slow, requiring a number of forward passes that grows with model size. From the many existing attribution methods \citep{nanda2023attribution,ferrando2024information,kramar2024atp}, we select edge attribution patching with integrated gradients (EAP-IG \citep{hanna2024circuit}) due to its faithful circuit-finding ability. Much like its predecessor, edge attribution patching (EAP \citep{nanda2023attribution}), EAP-IG assigns each edge an importance score using a gradient-based approximation of the change in loss that would occur if that edge were corrupted; however, EAP-IG yields more faithful circuits with fewer edges.

After running EAP-IG to score each edge, we define our circuit by greedily searching for the edges with the highest absolute score. We search for the minimal circuit that achieves at least 80\% of the whole model's performance on the task. We do this using binary search over circuit sizes; the initial search space ranges from 1 edge to 5\% of the model's edges. The high faithfulness threshold we set gives us confidence that our circuits capture most model mechanisms used on the given task. However, ensuring that a circuit is entirely complete, containing all relevant model nodes and edges, is challenging, and no definitive method of verifying this has emerged. The most notable existing method, from \citet{wang2023interpretability}, requires comparing circuit and model performance under a wide variety of ablations, and is seldom used due to its complexity and computational cost.

\subsection{Models}\label{sec:models}
We study \citeposs{biderman2023pythia} Pythia model suite, a collection of open-source autoregressive language models that includes intermediate training checkpoints. 
Though we could train our own language models or use another model suite with intermediate checkpoints \citep{sellam2022the,liu2023llm360,groeneveld2024olmo}, Pythia is unique in providing checkpoints for models at a variety of scales and training configurations.\footnote{We exclude OLMo from our analysis due to missing checkpoints at the time of writing.} Each model in the Pythia suite has 154 checkpoints: 11 of these correspond to the model after 0, 1, 2, 4, \ldots, and 512 steps of training; the remaining 143 correspond to 1000, 2000, \ldots, and 143,000 steps. We find circuits at each of these checkpoints. As Pythia uses a uniform batch size of 2.1 million tokens, these models are trained on far more tokens (300 billion) than those in existing studies of model internals over time. We study models of varying sizes, from 70 million to 12 billion parameters.  

\subsection{Tasks}\label{sec:tasks}
We analyze the mechanisms behind four different tasks taken from the (mechanistic) interpretability literature. We choose these tasks because they are simple and feasible for even the smaller models we study. Moreover, as existing work has already studied them in other models, we have clues as to how our models likely perform these tasks; to verify that our models use similar circuits we briefly analyze our models' indirect object identification and greater-than circuits in \Cref{app:task-circuits}. The other task are MLP-dominant and do not involve much attention head activity; for these circuits, we verify that this is still the case in Pythia models.

\paragraph{Indirect Object Identification} The indirect object identification (IOI, \citep{wang2023interpretability}) task feeds models inputs such as ``When John and Mary went to the store, John gave a drink to''; models should prefer \textit{Mary} over \textit{John}. Corrupted inputs, like ``When John and Mary went to the store, Mary gave a drink to'', reverse model preferences. We measure model behavior via the difference in logits assigned to the two names (\textit{Mary} and \textit{John}). We use a small dataset of 70 IOI examples created with \citeposs{wang2023interpretability} generator, as larger datasets did not provide significantly better results in our experiments and this size fit into GPU memory more easily.

\paragraph{Gendered-Pronoun} The Gendered-Pronoun task \citep{vig2020causal,mathwin2023identifying,chintam-etal-2023-identifying} measures the gender of the pronouns that models produce to refer to a previously mentioned entity. Prior work has shown ``So Paul is such a good cook, isn't"'', models prefer the continuation ``he'' to ``she''; we measure the degree to which this occurs via the difference in the pronouns' logits. In the corrupted case, we replace the ``Jack'' with ``Mary''; we include opposite-bias examples as well. We craft 70 examples as in \citep{mathwin2023identifying}.

\paragraph{Greater-Than} The Greater-Than task \citep{hanna2023how} measures a model's ability to complete inputs such as $s=$``The war lasted from the year 1732 to the year 17'' with a valid year (i.e. a year > 32). Task performance is measured via probability difference (prob diff); in this example, the prob diff is $\sum_{y=33}^{99} p(y|s) - \sum_{y=00}^{32}p(y|s)$. In corrupted inputs, the last two digits of the start year are replaced by ``01'', pushing the model to output early (invalid) years that decrease the prob diff. We create 200 Greater-Than examples with \citeposs{hanna2023how} generator.

\paragraph{Subject-Verb Agreement} Subject-verb agreement (SVA), widely studied within the NLP interpretability literature \cite{linzen-etal-2016-assessing,newman-etal-2021-refining,lasri-etal-2022-probing}, tasks models with predicting verb forms that match a sentence's subject. Given input such as ``The keys on the cabinet'', models must predict ``are'' over ``is''; a corrupted input, ``The key on the cabinet'' pushes models toward the opposite response. We measure model performance using prob diff, taking the difference of probability assigned to verbs that agree with the subject, and those that do not. We use 200 synthetic SVA example sentences from \citep{newman-etal-2021-refining}.

Note that two tasks (IOI and Gendered-Pronoun) use logit difference as their metric, while the other two (Greater-Than and SVA) use probability difference. In general, we prefer to compute differences of logits rather than probabilities, as the former may be better at detecting important negative components when performing patching \citep{zhang2024towards}. However, Greater-Than and SVA have multiple clean and corrupted answers, while logit difference, being unnormalized, is only compatible with one clean and corrupted answer. We thus compute a difference in probabilities instead, as in prior work.

\section{Circuit Formation}\label{sec:circuit-formation}
\subsection{Behavioral Evaluation}
\begin{figure}
    \centering
    \includegraphics[width=\textwidth]{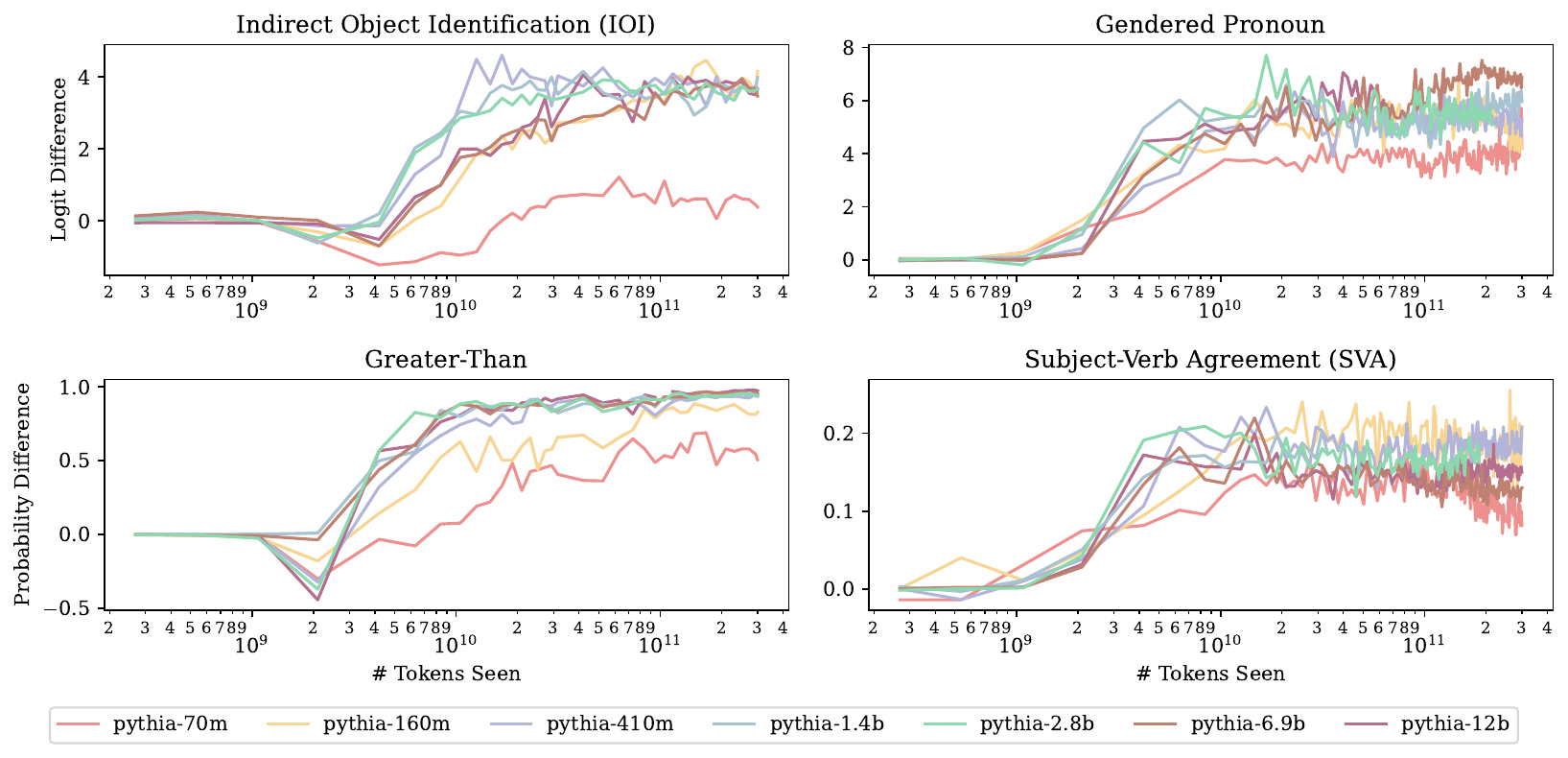}
    \caption{Task behavior across models and time (higher indicates a better match with expected behavior). Across tasks and scales, model abilities tend to develop at the same number of tokens.}
    \label{fig:behavioral-eval}
\end{figure}
We begin our analysis of LLMs' task mechanisms over time by analyzing LLM behavior on these tasks
; without understanding their task behaviors, we cannot understand their task mechanisms. We test these by running each model (\Cref{sec:models}) on each task (\Cref{sec:tasks}). Our results (\Cref{fig:behavioral-eval}) display three trends across all tasks. First, all models but the weakest (Pythia-70m) tend to arrive at the same task performance at the end of training. This is consistent with our choice of tasks: they are simple, learnable even by small models, and scaling does not significantly improve performance. Second, once models begin learning a task, their overall performance is generally non-decreasing, though there are minor fluctuations; Pythia-2.8b's logit difference on Gendered Pronouns dips slightly after it learns the task. In general, though, models tend not to undergo significant unlearning. The only marked downward trend (Pythia-70m at the end of SVA) comes from a weak model.

Finally, for each task we examined, we observed that there was a model size beyond which additional scale did not improve the rate of learning, and sometimes even decreased it; task acquisition appeared to approach an asymptote. We found this surprising due to the existence of findings showing the opposite trend for some tasks: \citep{kaplan2020scaling,rae2022scaling}. On some tasks (Gendered Pronouns and Greater-Than), all models above a certain size (70M parameters for Gendered Pronouns and 160M for Greater-Than) learn tasks at roughly the same rate. On IOI, models from 410M to 2.8B parameters learn the task the fastest, but larger models (6.9B and 12B) have learning curves more like Pythia-160m. We obtain similar results on more difficult tasks like SciQ \citep{welbl-etal-2017-crowdsourcing}; for these results, see \Cref{app:learning-ceiling-evidence}. 


What drives this last trend, limiting how fast even large models learn tasks? To understand this, we must delve into the internal model components that support these behaviors and trends.

\subsection{Component Emergence}
Prior work \citep{olsson2022context,chen2024sudden,singh2024needs} has shown how a model's ability to perform a specific task can hinge on the development of certain components, i.e. the emergence of attention heads or MLPs with specific, task-beneficial behaviors. Prior work has also thoroughly characterized the components underlying model abilities two of our tasks, IOI and Greater-Than, at the end of training. We thus ask: is it the development of these components that causes the task learning trends we saw before? We focus on four main components, all of which are attention heads,\footnote{Though past work on e.g. Greater-Than studied task-relevant MLPs and neurons, what generalizable behavior they have is poorly understood, unlike the attention heads we study.} which we briefly describe here: 

\textbf{Induction Heads} \citep{olsson2022context} activate on sequences of the form \texttt{[A][B]\ldots[A]}, attending to and upweighting \texttt{[B]}. This allow models to recreate patterns in their input, and supports IOI and Greater-Than.

\textbf{Successor Heads} \citep{gould2023successor} identify sequential values in the input (e.g. ``11'' or ``Thursday'') and upweight their successor (e.g. ``12'' or ``Friday''); this supports Greater-Than behavior.

\textbf{Copy Suppression Heads} \citep{mcdougall2023copy} attend to previous words in the input, lowering the output probability of repeated tokens that are highly predicted in the residual stream input to the head. In the original IOI circuit, copy suppression heads hurt performance, downweighting the correct name. In contrast, we find (\Cref{app:component-metrics}) that they contribute positively to the Pythia IOI circuit by downweighting the incorrect name; this is possible because both names are already highly predicted in the input to these heads, and they respond by downweighting the most repeated one.

\textbf{Name-Mover Heads} \citep{wang2023interpretability} perform the last step of the IOI task, by attending to and copying the correct name. Unlike the other heads described so far, this behavior is specific to IOI-type tasks; their behavior across the entire data distribution has not yet been characterized.

\begin{figure}
    \centering
    \includegraphics[width=\textwidth]{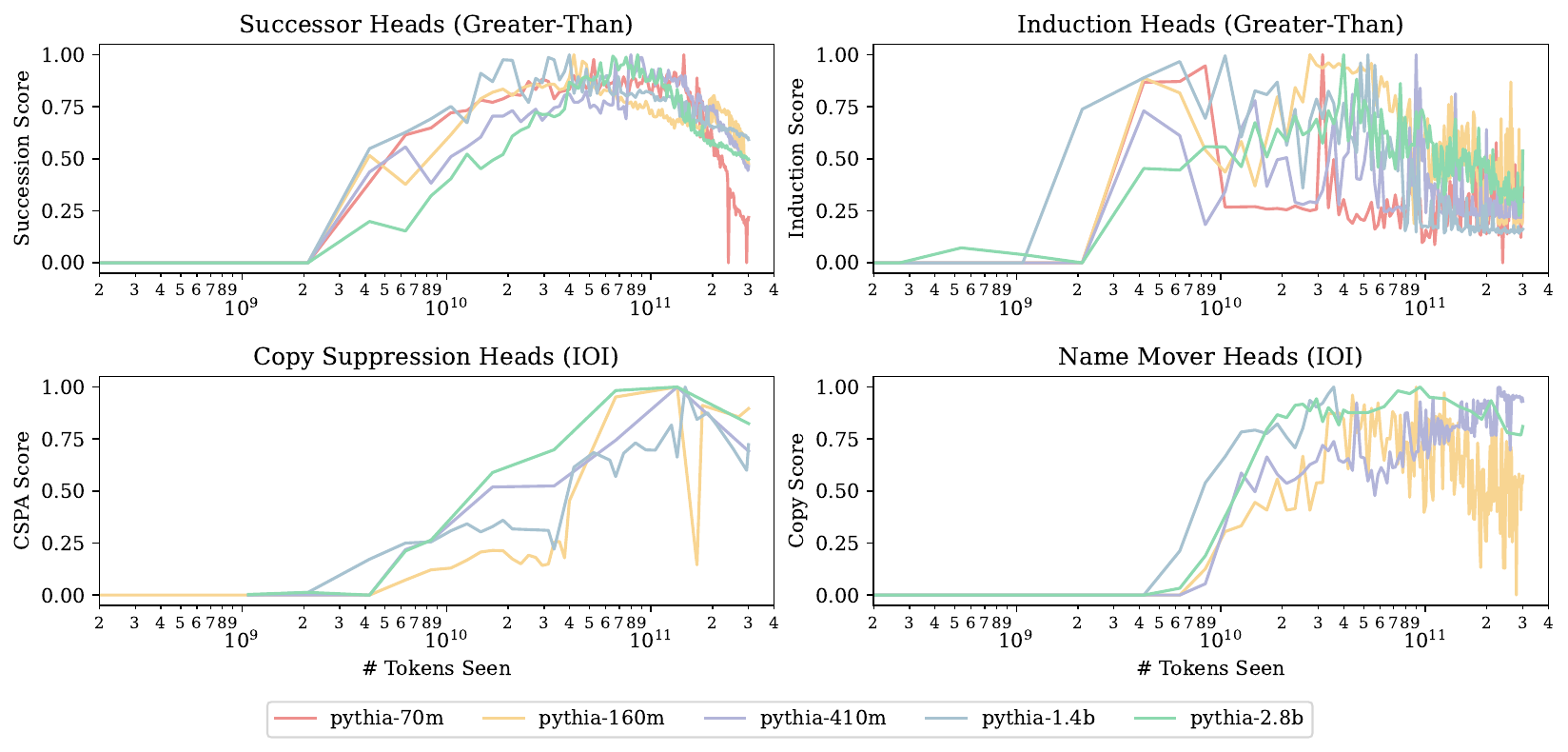}
    \caption{The development of components relevant to IOI and Greater-Than, across models and time. Each line indicates the degree to which attention heads in the circuit at each timestep exhibit the relevant component behavior. The timesteps at which component behavior emerges parallel those at which task performance emerges in \Cref{fig:behavioral-eval}.}
    \label{fig:component-eval}
\end{figure}

Because the importance of these components to IOI and Greater-Than has been established in other models, but not in necessarily in those of the Pythia suite, we must first confirm their importance in these models. To do so, we find circuits for each model at each checkpoint using EAP-IG, as described in \Cref{sec:circuit-finding}; note that circuits found at early checkpoints, where task performance is poor, are generally not meaningful. We omit Pythia-6.9b and 12b from circuit finding for reasons of computational cost. We then verify via path patching \citep{goldowskydill2023localizing} that these component types appear within Pythia models' tasks circuits; see \Cref{app:task-circuits,app:manual-circuit-analysis} for details on our methods and findings.

For each component, prior work has developed a metric to determine whether a model's attention head is acting like that component type; see \Cref{app:component-metrics} for details on these. Using these metrics, we score each of the heads in our models' circuits at each checkpoint, evaluating the degree to which each in-circuit head exhibits the aforementioned component behaviors. We then sum the score for each component across all in-circuit heads at each checkpoint, measuring how strongly the component behavior exists overall in the circuit at that checkpoint. We then normalize this sum across checkpoints, to understand how this behavior develops over time.

Our results (\Cref{fig:component-eval}) indicate that many of the hypothesized responsible components do emerge the same time as model performance increases. Most models' induction heads emerge soon after they have seen $2\times10^9$ tokens, replicating the findings in \citep{olsson2022context}; immediately after this, Greater-Than behavior emerges. The successor heads, also involved in Greater-Than, emerge in a similar timeframe. Curiously, the behavior of these heads eventually decreases, while task performance does not. We note that this could happen if the number of heads exhibiting successor behavior decreases, while the impact of said heads (on the model's logits) increases.

For IOI, the name-mover heads emerge at similar timesteps (2 - $8\times10^9$ tokens) across models, with a very high strength, during or just before IOI behavior appears. Copy suppression heads emerge at the same timescale, but at varying speeds, and with varying strengths. Given that these heads are the main contributors to model performance in each task's circuit, and they emerge as or just before models' task performance increases, we can be reasonably sure that they are responsible for the emergence of performance. The apparent ceiling on the speed at which these heads emerge, even for larger models, may be responsible for the same ceiling on how fast model performance emerges, underlain by these heads.

Throughout this section's analyses, we note an unusual trend: though model performance (\Cref{fig:behavioral-eval}) does not decrease over time, the functional behavior of certain attention heads does. In the following section, we explore how this occurs.

\section{Algorithmic Stability and Generalizability in Post-Formation Circuits}\label{sec:circuit-evolution}
We demonstrated in \Cref{sec:circuit-formation} that across a variety of tasks, models with differing sizes learn to perform the given task after the same amount of training; this appears to happen because each task relies on a set of components which develop after a similar count of training tokens across models. However, in \Cref{fig:component-eval}, we observed that attention heads that performed a given functional behavior at a given point in training may later perform that behavior more weakly or not at all. This raises questions: when the heads being used to solve a task change, does the algorithm implemented by the model change too? And how do these algorithms generalize across model scale?

\begin{figure}
    \centering
    \includegraphics[width=\textwidth]{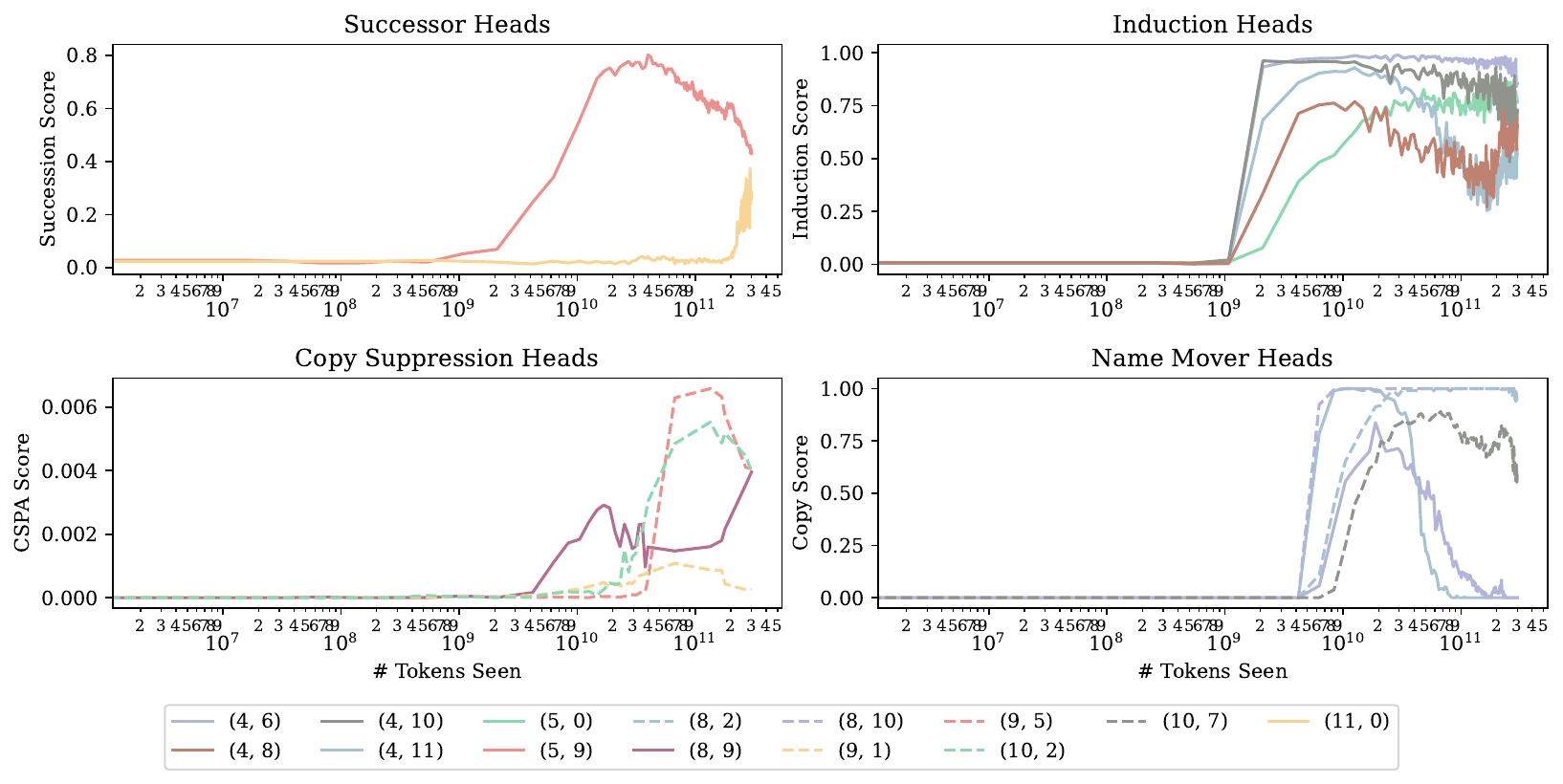}
    \caption{The development over time of components relevant to IOI and Greater-Than in Pythia-160m. Each line indicates the degree to which an attention head, denoted as (layer, head), exhibits a given function; higher values imply stronger functional behavior. Heads often lose their current function; as this occurs, other heads take their place.}
    \label{fig:component-eval-all-training}
\end{figure}

\subsection{Model Behavior and Circuit Components Post-Formation}
To understand how model components change over time, we analyze components and their algorithmic roles across training. We now zoom in on the components in one model, Pythia-160m, and study them over the course of training; where we earlier plotted only the top component (e.g. the top successor head), of each model, we now plot the top 5 of Pythia-160m's heads that exhibit a given functional behavior (or fewer, if fewer than 5 exist). By evaluating components and algorithms over Pythia-160m's 300B token training span, we extend beyond previous work, which studies models trained on relatively few ($\leq 50$M) tokens \citep{chen2024sudden, singh2024needs}; in such work, components and task behaviors appear constant after component formation.

By contrast, our results (\Cref{fig:component-eval-all-training}) show that over the longer training period of Pythia models, the identity of components in each circuit is not constant. For example, the name-mover head (4,6) suddenly stops exhibiting this behavior at $3\times10^{10}$ tokens, having acquired it after $4\times10^9$ tokens. Similarly, Pythia-160m's main successor head (5,9) loses its successor behavior towards the end of training; however, (11,0) exhibits more successor behavior at precisely that time. Such balancing may lead to the model's  task performance remaining stable, as we observed in the prior section (\Cref{fig:behavioral-eval}). It seems plausible that self-repair \citep{mcgrath2023hydra,rushing2024explorations} contributes to this behavioral stability, but we leave the question of the exact ``load-balancing'' mechanism to future work. Nevertheless, models can clearly compensate for losses of and changes in individual circuit components.

\subsection{Circuit Algorithm Stability Over Training} \label{subsec:algorithm-stability-over-training}
This instability of functional components raises an important question---when attention heads begin or cease to participate in a circuit, does the underlying algorithm change? To answer this, we examined the IOI circuit, as it is the most thoroughly characterized \citep{wang2023interpretability} circuit algorithm of our set of tasks. Our investigation follows a three-stage approach: first, we analyzed the IOI circuit at the end of training, reverse-engineering its algorithm; next, we developed a set of metrics to quantify whether the model was still performing that algorithm; finally, we applied these metrics across checkpoints, to determine if the algorithm was stable over training.

At each checkpoint, we identified relevant heads through two criteria:
\begin{enumerate}
    \item The head's effect on model performance when ablated via path patching
    \item The head's score on component-specific functional tests (e.g., copy suppression score for name-mover heads)
\end{enumerate}

Heads were included in our analysis only if they both: scored above a 10\% threshold on their respective functional tests, and showed a negative effect on the logit difference when ablated. For S2-inhibition heads, we additionally verified that ablating positional information while preserving token information reduced both logit difference and name-mover head attention to the indirect object while increasing attention to the subject.

\begin{figure}[ht]
    \centering
        \includegraphics[width=0.49\textwidth]{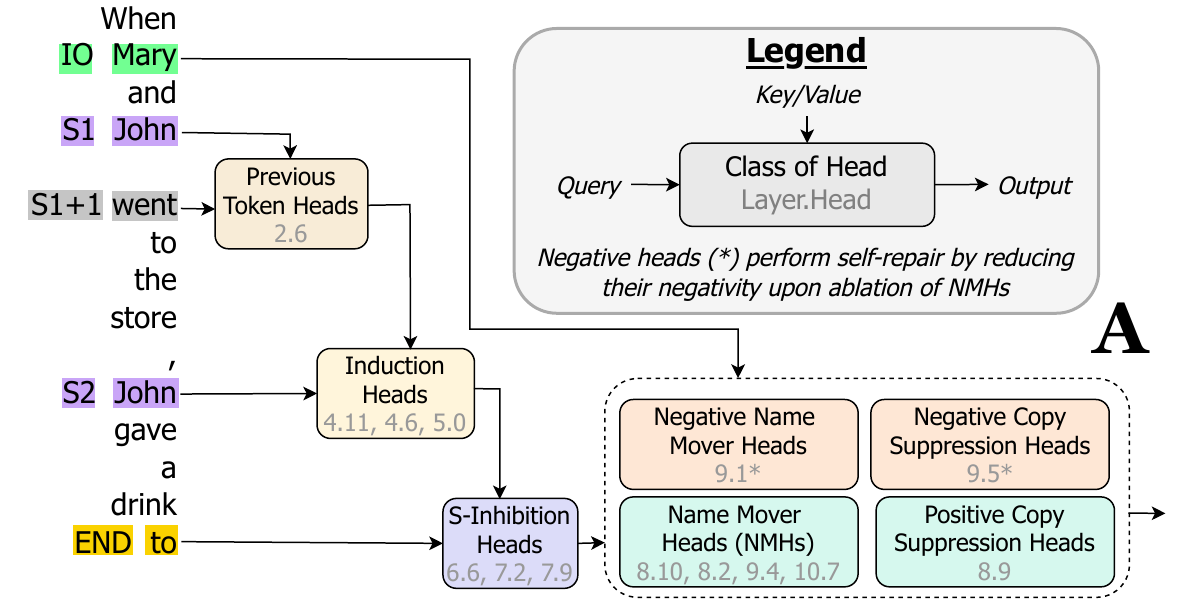}
        \includegraphics[width=0.49\textwidth]{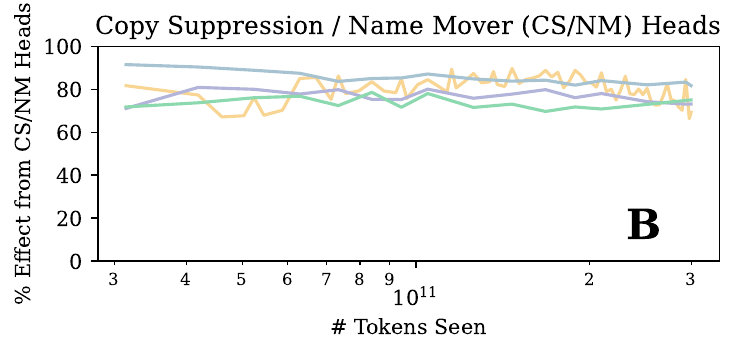}
        \includegraphics[width=\textwidth]{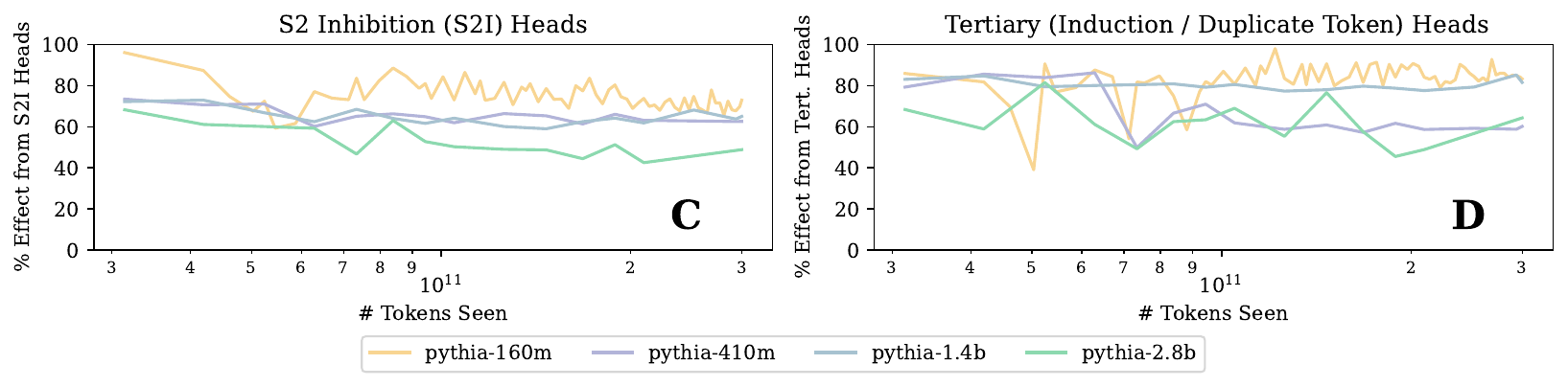}
    \caption{\textbf{A}: Pythia-160m's IOI circuit at the end of training (300B tokens). The remaining plots show the percent of model IOI performance that is explained by the Copy Suppression and Name-Mover Heads (\textbf{B}), the S-Inhibition Heads' edges to those heads (\textbf{C}), and the Induction / Duplicate Token Heads' connections to the S-Inhibition heads (\textbf{D}); higher percentages indicate that the corresponding edge is indeed important. Each of plots \textbf{B}-\textbf{D} verifies the importance of an edge from diagram \textbf{A}.}
    \label{fig:pythia-ioi-circuits}
\end{figure}

The first stage of our analysis is to analyze the IOI circuit at the end of training. Here, we present only the results of our analysis, but see \Cref{app:manual-circuit-analysis} for details of this process, which follows the original analysis \citep{wang2023interpretability}. \Cref{fig:pythia-ioi-circuits}\textbf{A} shows the circuit that results from our analysis; it involves three logical ``steps,'' each of which involves a different set of attention head types. Working backwards from the logit predictions, the direct contributors towards the logit difference are name-mover heads and copy suppression heads. The former attend to the indirect object in the prompt and copy it to the last position; the latter attend to and downweight tokens that appear earlier in the input. In the next step, the name-mover heads (but not the copy-suppression heads) use on token and positional information output by the S-inhibition heads to attend to the correct token. Finally, S-inhibition heads rely on information from induction heads and duplicate-token heads.

Next, we quantify the extent to which the circuit depends on each of these three steps, via path patching \citep{goldowskydill2023localizing}. When using path patching, we intervene on a node H (e.g., S-inhibition heads), route the change through an intermediate node or set of nodes R that depends on H (e.g., name-mover heads), and finally measure how the change in R affects the model output. For each step, our metric is the effect (on logit difference) of ablating the targeted components (through any intermediate nodes) divided by the effect of ablating all components that could affect R. For example, when measuring S-inhibition heads' importance to name-mover heads, the denominator is the ablation effect of all heads upstream of the name-mover heads. Higher ratios (0-100\%) indicate greater relative importance of the targeted components. More details on path patching can be found in \Cref{app:manual-circuit-analysis}, while details on the algorithm experiment itself can be found in \Cref{app:task-circuits}.

Finally, we compute each of these metrics for each model from 160M to 2.8B parameters in size.\footnote{We omit Pythia-70m, as it does not learn the task; due to computational constraints, we omit Pythia-6.9b/12b.} We run them on each checkpoint post-circuit emergence (that is, when all component types appear in the circuit); for Pythia-160m, we test every checkpoint, and for the larger models we space out checkpoints to save compute, using approximately 1/3rd of the available checkpoints). We find (\Cref{fig:pythia-ioi-circuits}\textbf{B}-\textbf{D}) that the behavior measured by these metrics is stable once the initial circuit has formed. Notably, in no model or metric are there dramatic shifts in algorithm corresponding to functional component shifts within the circuit. Moreover, all scores are relatively high, generally above 50\%; the core solvers of the algorithm, copy suppression and name-mover heads, have scores above 70\%. This suggests that analyses of circuits in fully pre-trained models may generalize well to other model states, rather than being contingent on the particular checkpoint selected.

Generalization across model scales also seems promising, as IOI circuit metrics from Pythia-160m are also high in larger Pythia variants. However, there is variation: while the name-mover, copy-suppression, and S-inhibition heads are at work in all models' circuits, the Pythia-160m circuit does not involve duplicate-token heads, while others do. So small differences exist amid big-picture similarity. Moreover, we stress that these algorithmic similarities might not hold for more complex tasks, for which a greater variety of algorithms could exist.

\section{Graph-Level Circuit Analysis} \label{sec:graph-level-analysis}
We have now examined circuits over the training process from component and algorithmic perspectives. But how do the circuit subgraphs themselves change over time and scale? In \Cref{sec:circuit-finding}, we explain how these subgraphs are collected; we applied this method to Pythia-70m through Pythia-2.8b for the tasks listed in \Cref{sec:tasks}. The result is a set of nodes and edges for each model, checkpoint, and task. Here, we briefly examine some trends we identified in the analysis of this data.

\begin{figure}[htb]
  \centering
  \includegraphics[width=\textwidth]{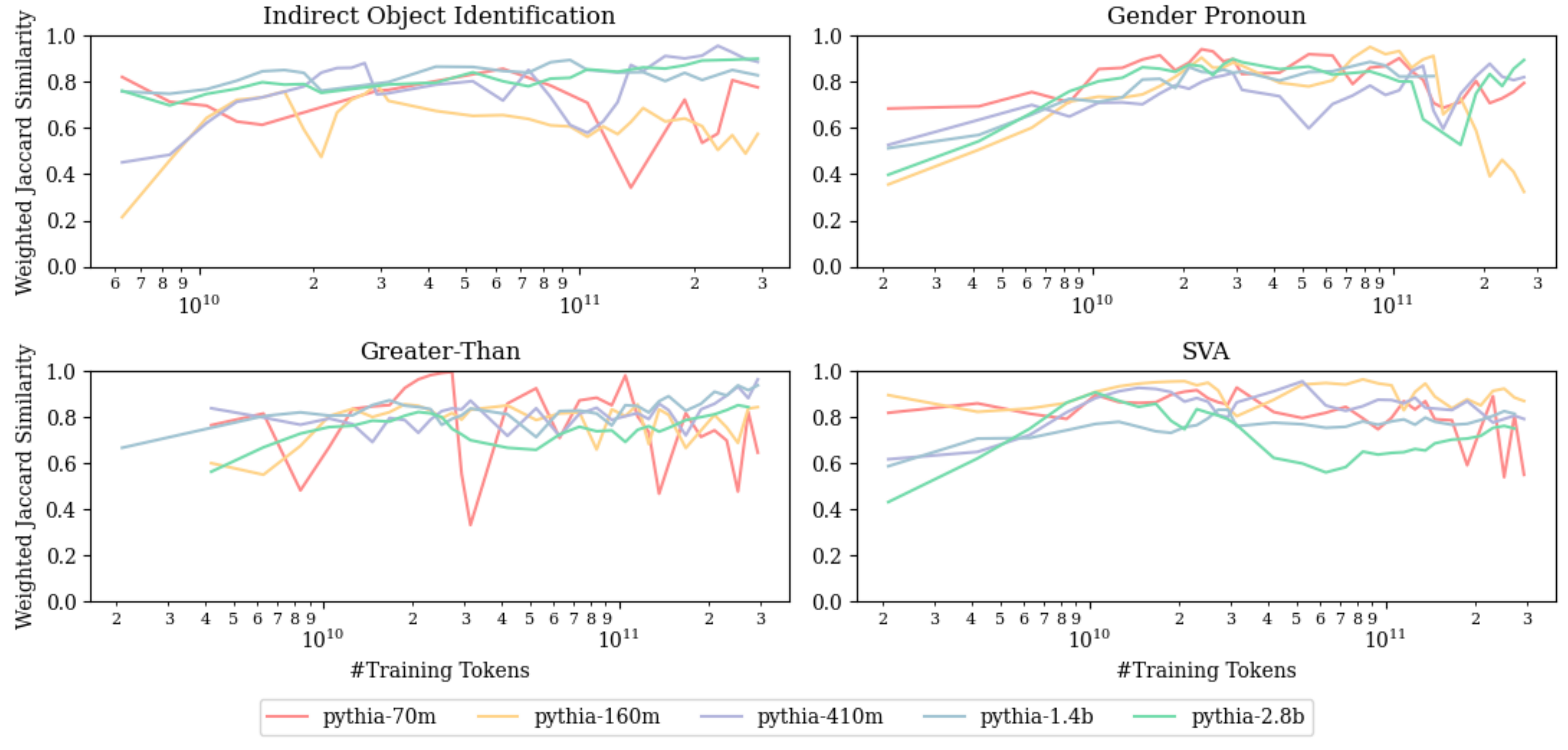} 
  \caption{Exponentially-weighted moving average Jaccard similarity for circuit node sets over training token count. In general, larger models tend to have both higher average EWMA-JS and fewer abrupt fluctuations, indicating higher stability in the circuit constituents.}
  \label{fig:jaccard similarity}
\end{figure}
We first examine the consistency of the nodes in circuits over training. To measure this, we compute the Jaccard similarity (intersection over union) $x_t$ between the circuits at each given checkpoint and those at all previous checkpoints. In order to smooth out local fluctuations and observe longer-term trends, we apply an exponential weighting with a decay factor $\alpha=0.5$, such that the value at a given checkpoint is the exponentially-weighted Jaccard similarity with the complete set of previous checkpoints. We calculate the weighted Jaccard similarity $\hat{x_t}$ at checkpoint $t$: $\hat{x_t} = 0.5 \hat{x_{t-1}} + 0.5 x_{t}$,
Our results (\Cref{fig:jaccard similarity}) suggest that larger models tend to form more stable circuits (with both higher average values and fewer sharp fluctuations); EWMA-Jaccard is more volatile for Pythia-70m/160m. In the Gendered-Pronoun circuit, we observe that significant changes can occur even late in training. We also computed how similar the circuit is during training with respect to the final circuit in Figure \ref{fig:njs}. we see more fluctuations, indicating that components swap during training at every checkpoint. Meanwhile, the upward trend indicates that circuits grow more and more similar to the final circuit during training.

We also compare the sizes (node counts) of the circuits over training. Across all four of our tasks, we find that the circuit size is positively correlated with the size of the models. We averaged node count across all the checkpoints for all the models on four tasks and calculated the pairwise correlation between the sizes of the models and the average node sizes. The Pearson correlation is $r=$0.72 for IOI and SVA, 0.9 for Greater-Than, 0.6 for Gender Pronoun.  We also find a high degree of variability---circuit sizes can remain stable or fluctuate significantly, with no clear pattern based on the model or the task. We leave further exploration of why this is the case to future work, but present our size metrics in \Cref{app:change-graphs}, \Cref{fig:circuit-size}.

\section{Discussion}
\paragraph{Implications for Interpretability Research}
While our findings are based on a limited set of circuits, they hold significant implications for mechanistic interpretability research. Our study was motivated by the fact that most such research does not study models that vary over time, like currently deployed models. However, the stability of circuit algorithms over the course of training suggests that analyses performed on models at a given point during training may provide valuable insights into earlier and later phases of training as well. Moreover, the consistency in the emergence of critical components and the algorithmic structure of these circuits across different model scales suggests that studying smaller models can sometimes provide insights applicable to larger models. This dual stability across training and scale could reduce the computational burden of interpretability research and allow for more efficient study of model mechanisms. However, further research is needed to confirm these trends across a broader range of tasks and model architectures.

\paragraph{Quantization Model of Neural Scaling}
Our findings suggest evidence for the quantization model of neural scaling  \citep{michaud2024quantizationmodelneuralscaling}, which suggests that models learn tasks in order of decreasing use frequency. This model may explain why we observed models learning tasks (and sub-tasks, represented by circuit functional components) at similar times despite model size. As all Pythia models are trained on the same dataset, the same distribution of data (and thus task frequency) would be presented to each model, leading to acquisition of those common tasks.

\paragraph{Limitations and Future Work} \label{subsec:limitations}
Our analysis was limited to a narrow range of tasks feasible for small models. This limits in turn the scope of the claims that we can make. We believe it to be very possible that more complex tasks, not solvable by small models, which permit a larger range of algorithmic solutions, may show different trends from those that we discuss here. Such work would be valuable, though computationally expensive due to the model sizes required. Our analysis also studied models only from one model family, Pythia. It is thus not possible to tell if our results are limited to the specific model family we have chosen, which shares both architecture and training setup across model scale. Such work is in part hampered by the lack of large-scale model suites such as Pythia; future work could provide these suites to enable this sort of analysis. Finally, future work would do well to explore more complex phenomena, such as the self-repair and load-balancing mechanisms of LLMs, which ensure consistent task performance despite component fluctuations.


\section{Related Work} \label{related_work}
\paragraph{Interpretability Over Time} LLMs' development over the course of pre-training has been studied with various non-mechanistic interpretability techniques, particularly behavioral interpretability, which characterizes model behavior without making claims about its implementation. Such longitudinal analyses have studied LLM learning curves and shown that models of different sizes acquire capabilities in the same sequence \citep{xia-etal-2023-training,chang2023characterizing}, examined how LLMs learn linguistic information \citep{warstadt-etal-2020-learning,choshen-etal-2022-grammar,chang-bergen-2022-word} and even predicted LLM behavior later in training \citep{hu2023latent,biderman2023emergent}. 
Nevertheless, behavioral studies alone cannot inform us about model internals. Prior work has studied the development of mechanisms in smaller models \citep{nanda2023progress,olsson2022context}, and suggests that model mechanisms can change abruptly, even as models' outward behavior stays the same. Other previous studies have examined the pre-training window where acquisition of extrinsic grammatical capabilities occurs \citep{chen2024sudden}.

\paragraph{Mechanistic Interpretability}
We build on previous work in mechanistic interpretability, which aims to reverse engineer neural networks. \textit{Circuits} are a significant paradigm of model analysis that has emerged from this field, originating with vision models \citep{olah2020zoom} and continuing to transformer LMs \citep{meng2023locating,wang2023interpretability,hanna2023how, varma2023explaining, merullo2024circuit, lieberum2023does, tigges2023linear}. Increasingly, research has tried to characterize the individual components at work within circuits, not only at the level of attention heads \citep{olsson2022context,chen2024sudden, singh2024needs, gould2023successor, mcdougall2023copy}, but also neurons \citep{vig2020causal,finlayson-etal-2021-causal,sajjad2022neuronlevel,gurnee2023finding,voita2023neurons} and other sorts of features \citep{bricken2023monosemanticity,huben2024sparse,marks2024sparse}. Recent work has also tried to accelerate mechanistic research via automated techniques \citep{conmy2023towards, bills2023language, syed2023attribution, hanna2024circuit}. Though mechanistic interpretability is a diverse field, it is often tied together by a reliance on causal methods \citep{vig2020causal, chan2022causal, Geiger:Lu-etal:2021, Geiger-etal:2023:CA, meng2023locating, wang2023interpretability, scrubbing, gevaFact}, which provide more faithful mechanistic explanations. 

\section*{Acknowledgements}
MH is supported in part by an OpenAI Superalignment Fellowship.

\bibliography{references}

\begin{thebibliography}{75}
\providecommand{\natexlab}[1]{#1}
\providecommand{\url}[1]{\texttt{#1}}
\expandafter\ifx\csname urlstyle\endcsname\relax
  \providecommand{\doi}[1]{doi: #1}\else
  \providecommand{\doi}{doi: \begingroup \urlstyle{rm}\Url}\fi

\bibitem[Adebayo et~al.(2018)Adebayo, Gilmer, Muelly, Goodfellow, Hardt, and Kim]{Adebayo2018SanityCF}
Julius Adebayo, Justin Gilmer, Michael Muelly, Ian~J. Goodfellow, Moritz Hardt, and Been Kim.
\newblock Sanity checks for saliency maps.
\newblock In \emph{Neural Information Processing Systems}, 2018.
\newblock URL \url{https://api.semanticscholar.org/CorpusID:52938797}.

\bibitem[Anthropic(2024)]{anthropic2024claude}
Anthropic.
\newblock The claude 3 model family: Opus, sonnet, haiku, 2024.
\newblock URL \url{https://www-cdn.anthropic.com/de8ba9b01c9ab7cbabf5c33b80b7bbc618857627/Model_Card_Claude_3.pdf}.

\bibitem[Belinkov(2022)]{belinkov-2022-probing}
Yonatan Belinkov.
\newblock Probing classifiers: Promises, shortcomings, and advances.
\newblock \emph{Computational Linguistics}, 48\penalty0 (1):\penalty0 207--219, March 2022.
\newblock \doi{10.1162/coli_a_00422}.
\newblock URL \url{https://aclanthology.org/2022.cl-1.7}.

\bibitem[Biderman et~al.(2023{\natexlab{a}})Biderman, Prashanth, Sutawika, Schoelkopf, Anthony, Purohit, and Raff]{biderman2023emergent}
Stella Biderman, USVSN~Sai Prashanth, Lintang Sutawika, Hailey Schoelkopf, Quentin~Gregory Anthony, Shivanshu Purohit, and Edward Raff.
\newblock Emergent and predictable memorization in large language models.
\newblock In \emph{Thirty-seventh Conference on Neural Information Processing Systems}, 2023{\natexlab{a}}.
\newblock URL \url{https://openreview.net/forum?id=Iq0DvhB4Kf}.

\bibitem[Biderman et~al.(2023{\natexlab{b}})Biderman, Schoelkopf, Anthony, Bradley, O'Brien, Hallahan, Khan, Purohit, Prashanth, Raff, Skowron, Sutawika, and Van Der~Wal]{biderman2023pythia}
Stella Biderman, Hailey Schoelkopf, Quentin Anthony, Herbie Bradley, Kyle O'Brien, Eric Hallahan, Mohammad~Aflah Khan, Shivanshu Purohit, USVSN~Sai Prashanth, Edward Raff, Aviya Skowron, Lintang Sutawika, and Oskar Van Der~Wal.
\newblock Pythia: a suite for analyzing large language models across training and scaling.
\newblock In \emph{Proceedings of the 40th International Conference on Machine Learning}, ICML'23. JMLR.org, 2023{\natexlab{b}}.

\bibitem[Bills et~al.(2023)Bills, Cammarata, Mossing, Tillman, Gao, Goh, Sutskever, Leike, Wu, and Saunders]{bills2023language}
Steven Bills, Nick Cammarata, Dan Mossing, Henk Tillman, Leo Gao, Gabriel Goh, Ilya Sutskever, Jan Leike, Jeff Wu, and William Saunders.
\newblock Language models can explain neurons in language models.
\newblock \url{https://openaipublic.blob.core.windows.net/neuron-explainer/paper/index.html}, 2023.

\bibitem[Bilodeau et~al.(2024)Bilodeau, Jaques, Koh, and Kim]{bilodeau2024impossibility}
Blair Bilodeau, Natasha Jaques, Pang~Wei Koh, and Been Kim.
\newblock Impossibility theorems for feature attribution.
\newblock \emph{Proceedings of the National Academy of Sciences}, 121\penalty0 (2):\penalty0 e2304406120, 2024.
\newblock \doi{10.1073/pnas.2304406120}.
\newblock URL \url{https://www.pnas.org/doi/abs/10.1073/pnas.2304406120}.

\bibitem[Bisk et~al.(2019)Bisk, Zellers, Bras, Gao, and Choi]{piqa}
Yonatan Bisk, Rowan Zellers, Ronan~Le Bras, Jianfeng Gao, and Yejin Choi.
\newblock Piqa: Reasoning about physical commonsense in natural language, 2019.

\bibitem[Bricken et~al.(2023)Bricken, Templeton, Batson, Chen, Jermyn, Conerly, Turner, Anil, Denison, Askell, Lasenby, Wu, Kravec, Schiefer, Maxwell, Joseph, Hatfield-Dodds, Tamkin, Nguyen, McLean, Burke, Hume, Carter, Henighan, and Olah]{bricken2023monosemanticity}
Trenton Bricken, Adly Templeton, Joshua Batson, Brian Chen, Adam Jermyn, Tom Conerly, Nick Turner, Cem Anil, Carson Denison, Amanda Askell, Robert Lasenby, Yifan Wu, Shauna Kravec, Nicholas Schiefer, Tim Maxwell, Nicholas Joseph, Zac Hatfield-Dodds, Alex Tamkin, Karina Nguyen, Brayden McLean, Josiah~E Burke, Tristan Hume, Shan Carter, Tom Henighan, and Christopher Olah.
\newblock Towards monosemanticity: Decomposing language models with dictionary learning.
\newblock \emph{Transformer Circuits Thread}, 2023.
\newblock https://transformer-circuits.pub/2023/monosemantic-features/index.html.

\bibitem[Chan et~al.(2022)Chan, Garriga-Alonso, Goldwosky-Dill, Greenblatt, Nitishinskaya, Radhakrishnan, Shlegeris, and Thomas]{chan2022causal}
Lawrence Chan, Adrià Garriga-Alonso, Nicholas Goldwosky-Dill, Ryan Greenblatt, Jenny Nitishinskaya, Ansh Radhakrishnan, Buck Shlegeris, and Nate Thomas.
\newblock Causal scrubbing, a method for rigorously testing interpretability hypotheses.
\newblock \emph{AI Alignment Forum}, 2022.
\newblock \url{https://www.alignmentforum.org/posts/JvZhhzycHu2Yd57RN/causal-scrubbing-a-method-for-rigorously-testing}.

\bibitem[Chan et~al.(2023)Chan, Garriga-Alonso, Goldowsky-Dill, Greenblatt, Nitishinskaya, Radhakrishnan, Shlegeris, and Thomas]{scrubbing}
Lawrence Chan, Adrià Garriga-Alonso, Nicholas Goldowsky-Dill, Ryan Greenblatt, Jenny Nitishinskaya, Ansh Radhakrishnan, Buck Shlegeris, and Nate Thomas.
\newblock Causal scrubbing: a method for rigorously testing interpretability hypotheses [redwood research].
\newblock Alignment Forum, 2023.
\newblock URL \url{https://www.alignmentforum.org/posts/JvZhhzycHu2Yd57RN/causal-scrubbing-a-method-for-rigorously-testing}.
\newblock Accessed: 17th Sep 2023.

\bibitem[Chang and Bergen(2022)]{chang-bergen-2022-word}
Tyler~A. Chang and Benjamin~K. Bergen.
\newblock Word acquisition in neural language models.
\newblock \emph{Transactions of the Association for Computational Linguistics}, 10:\penalty0 1--16, 2022.
\newblock \doi{10.1162/tacl_a_00444}.
\newblock URL \url{https://aclanthology.org/2022.tacl-1.1}.

\bibitem[Chang et~al.(2023)Chang, Tu, and Bergen]{chang2023characterizing}
Tyler~A. Chang, Zhuowen Tu, and Benjamin~K. Bergen.
\newblock Characterizing learning curves during language model pre-training: Learning, forgetting, and stability, 2023.

\bibitem[Chen et~al.(2024)Chen, Shwartz-Ziv, Cho, Leavitt, and Saphra]{chen2024sudden}
Angelica Chen, Ravid Shwartz-Ziv, Kyunghyun Cho, Matthew~L Leavitt, and Naomi Saphra.
\newblock Sudden drops in the loss: Syntax acquisition, phase transitions, and simplicity bias in {MLM}s.
\newblock In \emph{The Twelfth International Conference on Learning Representations}, 2024.
\newblock URL \url{https://openreview.net/forum?id=MO5PiKHELW}.

\bibitem[Chintam et~al.(2023)Chintam, Beloch, Zuidema, Hanna, and van~der Wal]{chintam-etal-2023-identifying}
Abhijith Chintam, Rahel Beloch, Willem Zuidema, Michael Hanna, and Oskar van~der Wal.
\newblock Identifying and adapting transformer-components responsible for gender bias in an {E}nglish language model.
\newblock In Yonatan Belinkov, Sophie Hao, Jaap Jumelet, Najoung Kim, Arya McCarthy, and Hosein Mohebbi, editors, \emph{Proceedings of the 6th BlackboxNLP Workshop: Analyzing and Interpreting Neural Networks for NLP}, pages 379--394, Singapore, December 2023. Association for Computational Linguistics.
\newblock \doi{10.18653/v1/2023.blackboxnlp-1.29}.
\newblock URL \url{https://aclanthology.org/2023.blackboxnlp-1.29}.

\bibitem[Choshen et~al.(2022)Choshen, Hacohen, Weinshall, and Abend]{choshen-etal-2022-grammar}
Leshem Choshen, Guy Hacohen, Daphna Weinshall, and Omri Abend.
\newblock The grammar-learning trajectories of neural language models.
\newblock In Smaranda Muresan, Preslav Nakov, and Aline Villavicencio, editors, \emph{Proceedings of the 60th Annual Meeting of the Association for Computational Linguistics (Volume 1: Long Papers)}, pages 8281--8297, Dublin, Ireland, May 2022. Association for Computational Linguistics.
\newblock \doi{10.18653/v1/2022.acl-long.568}.
\newblock URL \url{https://aclanthology.org/2022.acl-long.568}.

\bibitem[Chung et~al.(2022)Chung, Hou, Longpre, Zoph, Tay, Fedus, Li, Wang, Dehghani, Brahma, Webson, Gu, Dai, Suzgun, Chen, Chowdhery, Castro-Ros, Pellat, Robinson, Valter, Narang, Mishra, Yu, Zhao, Huang, Dai, Yu, Petrov, Chi, Dean, Devlin, Roberts, Zhou, Le, and Wei]{chung2022scaling}
Hyung~Won Chung, Le~Hou, Shayne Longpre, Barret Zoph, Yi~Tay, William Fedus, Yunxuan Li, Xuezhi Wang, Mostafa Dehghani, Siddhartha Brahma, Albert Webson, Shixiang~Shane Gu, Zhuyun Dai, Mirac Suzgun, Xinyun Chen, Aakanksha Chowdhery, Alex Castro-Ros, Marie Pellat, Kevin Robinson, Dasha Valter, Sharan Narang, Gaurav Mishra, Adams Yu, Vincent Zhao, Yanping Huang, Andrew Dai, Hongkun Yu, Slav Petrov, Ed~H. Chi, Jeff Dean, Jacob Devlin, Adam Roberts, Denny Zhou, Quoc~V. Le, and Jason Wei.
\newblock Scaling instruction-finetuned language models, 2022.

\bibitem[Clark et~al.(2018)Clark, Cowhey, Etzioni, Khot, Sabharwal, Schoenick, and Tafjord]{arc}
Peter Clark, Isaac Cowhey, Oren Etzioni, Tushar Khot, Ashish Sabharwal, Carissa Schoenick, and Oyvind Tafjord.
\newblock Think you have solved question answering? try arc, the ai2 reasoning challenge, 2018.

\bibitem[Cohen et~al.(2023)Cohen, Biran, Yoran, Globerson, and Geva]{gevaFact}
Roi Cohen, Eden Biran, Ori Yoran, Amir Globerson, and Mor Geva.
\newblock Evaluating the ripple effects of knowledge editing in language models, 2023.

\bibitem[Conmy et~al.(2023)Conmy, Mavor-Parker, Lynch, Heimersheim, and Garriga-Alonso]{conmy2023towards}
Arthur Conmy, Augustine~N. Mavor-Parker, Aengus Lynch, Stefan Heimersheim, and Adri{\`a} Garriga-Alonso.
\newblock Towards automated circuit discovery for mechanistic interpretability.
\newblock In \emph{Thirty-seventh Conference on Neural Information Processing Systems}, 2023.
\newblock URL \url{https://openreview.net/forum?id=89ia77nZ8u}.

\bibitem[Elazar et~al.(2020)Elazar, Ravfogel, Jacovi, and Goldberg]{Elazar2020AmnesicPB}
Yanai Elazar, Shauli Ravfogel, Alon Jacovi, and Yoav Goldberg.
\newblock Amnesic probing: Behavioral explanation with amnesic counterfactuals.
\newblock \emph{Transactions of the Association for Computational Linguistics}, 9:\penalty0 160--175, 2020.
\newblock URL \url{https://api.semanticscholar.org/CorpusID:227408471}.

\bibitem[Elhage et~al.(2021)Elhage, Nanda, Olsson, Henighan, Joseph, Mann, Askell, Bai, Chen, Conerly, DasSarma, Drain, Ganguli, Hatfield-Dodds, Hernandez, Jones, Kernion, Lovitt, Ndousse, Amodei, Brown, Clark, Kaplan, McCandlish, and Olah]{elhage2021mathematical}
Nelson Elhage, Neel Nanda, Catherine Olsson, Tom Henighan, Nicholas Joseph, Ben Mann, Amanda Askell, Yuntao Bai, Anna Chen, Tom Conerly, Nova DasSarma, Dawn Drain, Deep Ganguli, Zac Hatfield-Dodds, Danny Hernandez, Andy Jones, Jackson Kernion, Liane Lovitt, Kamal Ndousse, Dario Amodei, Tom Brown, Jack Clark, Jared Kaplan, Sam McCandlish, and Chris Olah.
\newblock A mathematical framework for transformer circuits.
\newblock \emph{Transformer Circuits Thread}, 2021.
\newblock https://transformer-circuits.pub/2021/framework/index.html.

\bibitem[Ferrando and Voita(2024)]{ferrando2024information}
Javier Ferrando and Elena Voita.
\newblock Information flow routes: Automatically interpreting language models at scale, 2024.

\bibitem[Ferrando et~al.(2024)Ferrando, Sarti, Bisazza, and Costa-jussà]{ferrando2024primer}
Javier Ferrando, Gabriele Sarti, Arianna Bisazza, and Marta~R. Costa-jussà.
\newblock A primer on the inner workings of transformer-based language models, 2024.

\bibitem[Finlayson et~al.(2021)Finlayson, Mueller, Gehrmann, Shieber, Linzen, and Belinkov]{finlayson-etal-2021-causal}
Matthew Finlayson, Aaron Mueller, Sebastian Gehrmann, Stuart Shieber, Tal Linzen, and Yonatan Belinkov.
\newblock Causal analysis of syntactic agreement mechanisms in neural language models.
\newblock In Chengqing Zong, Fei Xia, Wenjie Li, and Roberto Navigli, editors, \emph{Proceedings of the 59th Annual Meeting of the Association for Computational Linguistics and the 11th International Joint Conference on Natural Language Processing (Volume 1: Long Papers)}, pages 1828--1843, Online, August 2021. Association for Computational Linguistics.
\newblock \doi{10.18653/v1/2021.acl-long.144}.
\newblock URL \url{https://aclanthology.org/2021.acl-long.144}.

\bibitem[Geiger et~al.(2021)Geiger, Lu, Icard, and Potts]{Geiger:Lu-etal:2021}
Atticus Geiger, Hanson Lu, Thomas Icard, and Christopher Potts.
\newblock Causal abstractions of neural networks.
\newblock In \emph{Advances in Neural Information Processing Systems}, volume~34, pages 9574--9586, 2021.
\newblock URL \url{https://papers.nips.cc/paper/2021/hash/4f5c422f4d49a5a807eda27434231040-Abstract.html}.

\bibitem[Geiger et~al.(2023)Geiger, Potts, and Icard]{Geiger-etal:2023:CA}
Atticus Geiger, Christopher Potts, and Thomas Icard.
\newblock Causal abstraction for faithful model interpretation.
\newblock Ms., Stanford University, 2023.
\newblock URL \url{https://arxiv.org/abs/2301.04709}.

\bibitem[Goldowsky-Dill et~al.(2023)Goldowsky-Dill, MacLeod, Sato, and Arora]{goldowskydill2023localizing}
Nicholas Goldowsky-Dill, Chris MacLeod, Lucas Sato, and Aryaman Arora.
\newblock Localizing model behavior with path patching, 2023.

\bibitem[Gould et~al.(2023)Gould, Ong, Ogden, and Conmy]{gould2023successor}
Rhys Gould, Euan Ong, George Ogden, and Arthur Conmy.
\newblock Successor heads: Recurring, interpretable attention heads in the wild, 2023.

\bibitem[Groeneveld et~al.(2024)Groeneveld, Beltagy, Walsh, Bhagia, Kinney, Tafjord, Jha, Ivison, Magnusson, Wang, Arora, Atkinson, Authur, Chandu, Cohan, Dumas, Elazar, Gu, Hessel, Khot, Merrill, Morrison, Muennighoff, Naik, Nam, Peters, Pyatkin, Ravichander, Schwenk, Shah, Smith, Strubell, Subramani, Wortsman, Dasigi, Lambert, Richardson, Zettlemoyer, Dodge, Lo, Soldaini, Smith, and Hajishirzi]{groeneveld2024olmo}
Dirk Groeneveld, Iz~Beltagy, Pete Walsh, Akshita Bhagia, Rodney Kinney, Oyvind Tafjord, Ananya~Harsh Jha, Hamish Ivison, Ian Magnusson, Yizhong Wang, Shane Arora, David Atkinson, Russell Authur, Khyathi~Raghavi Chandu, Arman Cohan, Jennifer Dumas, Yanai Elazar, Yuling Gu, Jack Hessel, Tushar Khot, William Merrill, Jacob Morrison, Niklas Muennighoff, Aakanksha Naik, Crystal Nam, Matthew~E. Peters, Valentina Pyatkin, Abhilasha Ravichander, Dustin Schwenk, Saurabh Shah, Will Smith, Emma Strubell, Nishant Subramani, Mitchell Wortsman, Pradeep Dasigi, Nathan Lambert, Kyle Richardson, Luke Zettlemoyer, Jesse Dodge, Kyle Lo, Luca Soldaini, Noah~A. Smith, and Hannaneh Hajishirzi.
\newblock Olmo: Accelerating the science of language models, 2024.

\bibitem[Gurnee et~al.(2023)Gurnee, Nanda, Pauly, Harvey, Troitskii, and Bertsimas]{gurnee2023finding}
Wes Gurnee, Neel Nanda, Matthew Pauly, Katherine Harvey, Dmitrii Troitskii, and Dimitris Bertsimas.
\newblock Finding neurons in a haystack: Case studies with sparse probing, 2023.

\bibitem[Hanna et~al.(2023)Hanna, Liu, and Variengien]{hanna2023how}
Michael Hanna, Ollie Liu, and Alexandre Variengien.
\newblock How does {GPT}-2 compute greater-than?: Interpreting mathematical abilities in a pre-trained language model.
\newblock In \emph{Thirty-seventh Conference on Neural Information Processing Systems}, 2023.
\newblock URL \url{https://openreview.net/forum?id=p4PckNQR8k}.

\bibitem[Hanna et~al.(2024)Hanna, Pezzelle, and Belinkov]{hanna2024circuit}
Michael Hanna, Sandro Pezzelle, and Yonatan Belinkov.
\newblock Have faith in faithfulness: Going beyond circuit overlap when finding model mechanisms, 2024.

\bibitem[Hu et~al.(2021)Hu, Shen, Wallis, Allen-Zhu, Li, Wang, Wang, and Chen]{hu2021lora}
Edward~J. Hu, Yelong Shen, Phillip Wallis, Zeyuan Allen-Zhu, Yuanzhi Li, Shean Wang, Lu~Wang, and Weizhu Chen.
\newblock Lora: Low-rank adaptation of large language models, 2021.

\bibitem[Hu et~al.(2023)Hu, Chen, Saphra, and Cho]{hu2023latent}
Michael~Y. Hu, Angelica Chen, Naomi Saphra, and Kyunghyun Cho.
\newblock Latent state models of training dynamics.
\newblock \emph{Transactions on Machine Learning Research}, 2023.
\newblock ISSN 2835-8856.
\newblock URL \url{https://openreview.net/forum?id=NE2xXWo0LF}.

\bibitem[Huben et~al.(2024)Huben, Cunningham, Smith, Ewart, and Sharkey]{huben2024sparse}
Robert Huben, Hoagy Cunningham, Logan~Riggs Smith, Aidan Ewart, and Lee Sharkey.
\newblock Sparse autoencoders find highly interpretable features in language models.
\newblock In \emph{The Twelfth International Conference on Learning Representations}, 2024.
\newblock URL \url{https://openreview.net/forum?id=F76bwRSLeK}.

\bibitem[Kaplan et~al.(2020)Kaplan, McCandlish, Henighan, Brown, Chess, Child, Gray, Radford, Wu, and Amodei]{kaplan2020scaling}
Jared Kaplan, Sam McCandlish, Tom Henighan, Tom~B. Brown, Benjamin Chess, Rewon Child, Scott Gray, Alec Radford, Jeffrey Wu, and Dario Amodei.
\newblock Scaling laws for neural language models, 2020.

\bibitem[Kramár et~al.(2024)Kramár, Lieberum, Shah, and Nanda]{kramar2024atp}
János Kramár, Tom Lieberum, Rohin Shah, and Neel Nanda.
\newblock Atp*: An efficient and scalable method for localizing llm behaviour to components, 2024.

\bibitem[Lasri et~al.(2022)Lasri, Pimentel, Lenci, Poibeau, and Cotterell]{lasri-etal-2022-probing}
Karim Lasri, Tiago Pimentel, Alessandro Lenci, Thierry Poibeau, and Ryan Cotterell.
\newblock Probing for the usage of grammatical number.
\newblock In Smaranda Muresan, Preslav Nakov, and Aline Villavicencio, editors, \emph{Proceedings of the 60th Annual Meeting of the Association for Computational Linguistics (Volume 1: Long Papers)}, pages 8818--8831, Dublin, Ireland, May 2022. Association for Computational Linguistics.
\newblock \doi{10.18653/v1/2022.acl-long.603}.
\newblock URL \url{https://aclanthology.org/2022.acl-long.603}.

\bibitem[Lieberum et~al.(2023)Lieberum, Rahtz, Kramár, Nanda, Irving, Shah, and Mikulik]{lieberum2023does}
Tom Lieberum, Matthew Rahtz, János Kramár, Neel Nanda, Geoffrey Irving, Rohin Shah, and Vladimir Mikulik.
\newblock Does circuit analysis interpretability scale? evidence from multiple choice capabilities in chinchilla, 2023.

\bibitem[Linzen et~al.(2016)Linzen, Dupoux, and Goldberg]{linzen-etal-2016-assessing}
Tal Linzen, Emmanuel Dupoux, and Yoav Goldberg.
\newblock Assessing the ability of {LSTM}s to learn syntax-sensitive dependencies.
\newblock \emph{Transactions of the Association for Computational Linguistics}, 4:\penalty0 521--535, 2016.
\newblock \doi{10.1162/tacl_a_00115}.
\newblock URL \url{https://aclanthology.org/Q16-1037}.

\bibitem[Liu et~al.(2023)Liu, Qiao, Neiswanger, Wang, Tan, Tao, Li, Wang, Sun, Pangarkar, Fan, Gu, Miller, Zhuang, He, Li, Koto, Tang, Ranjan, Shen, Ren, Iriondo, Mu, Hu, Schulze, Nakov, Baldwin, and Xing]{liu2023llm360}
Zhengzhong Liu, Aurick Qiao, Willie Neiswanger, Hongyi Wang, Bowen Tan, Tianhua Tao, Junbo Li, Yuqi Wang, Suqi Sun, Omkar Pangarkar, Richard Fan, Yi~Gu, Victor Miller, Yonghao Zhuang, Guowei He, Haonan Li, Fajri Koto, Liping Tang, Nikhil Ranjan, Zhiqiang Shen, Xuguang Ren, Roberto Iriondo, Cun Mu, Zhiting Hu, Mark Schulze, Preslav Nakov, Tim Baldwin, and Eric~P. Xing.
\newblock Llm360: Towards fully transparent open-source llms, 2023.

\bibitem[Marks et~al.(2024)Marks, Rager, Michaud, Belinkov, Bau, and Mueller]{marks2024sparse}
Samuel Marks, Can Rager, Eric~J. Michaud, Yonatan Belinkov, David Bau, and Aaron Mueller.
\newblock Sparse feature circuits: Discovering and editing interpretable causal graphs in language models, 2024.

\bibitem[Mathwin et~al.(2023)Mathwin, Corlouer, Kran, Barez, and Nanda]{mathwin2023identifying}
Chris Mathwin, Guillaume Corlouer, Esben Kran, Fazl Barez, and Neel Nanda.
\newblock Identifying a preliminary circuit for predicting gendered pronouns in gpt-2 small, 2023.
\newblock URL \url{https://itch.io/jam/mechint/rate/1889871}.

\bibitem[McDougall et~al.(2023)McDougall, Conmy, Rushing, McGrath, and Nanda]{mcdougall2023copy}
Callum McDougall, Arthur Conmy, Cody Rushing, Thomas McGrath, and Neel Nanda.
\newblock Copy suppression: Comprehensively understanding an attention head, 2023.

\bibitem[McGrath et~al.(2023)McGrath, Rahtz, Kramar, Mikulik, and Legg]{mcgrath2023hydra}
Thomas McGrath, Matthew Rahtz, Janos Kramar, Vladimir Mikulik, and Shane Legg.
\newblock The hydra effect: Emergent self-repair in language model computations, 2023.

\bibitem[Meng et~al.(2023)Meng, Bau, Andonian, and Belinkov]{meng2023locating}
Kevin Meng, David Bau, Alex Andonian, and Yonatan Belinkov.
\newblock Locating and editing factual associations in gpt, 2023.

\bibitem[Merullo et~al.(2024)Merullo, Eickhoff, and Pavlick]{merullo2024circuit}
Jack Merullo, Carsten Eickhoff, and Ellie Pavlick.
\newblock Circuit component reuse across tasks in transformer language models.
\newblock In \emph{The Twelfth International Conference on Learning Representations}, 2024.
\newblock URL \url{https://openreview.net/forum?id=fpoAYV6Wsk}.

\bibitem[Michaud et~al.(2024)Michaud, Liu, Girit, and Tegmark]{michaud2024quantizationmodelneuralscaling}
Eric~J. Michaud, Ziming Liu, Uzay Girit, and Max Tegmark.
\newblock The quantization model of neural scaling, 2024.
\newblock URL \url{https://arxiv.org/abs/2303.13506}.

\bibitem[Nanda(2023)]{nanda2023attribution}
Neel Nanda.
\newblock Attribution {Patching}: {Activation} {Patching} {At} {Industrial} {Scale}, 2023.
\newblock URL \url{https://www.neelnanda.io/mechanistic-interpretability/attribution-patching}.

\bibitem[Nanda et~al.(2023)Nanda, Chan, Lieberum, Smith, and Steinhardt]{nanda2023progress}
Neel Nanda, Lawrence Chan, Tom Lieberum, Jess Smith, and Jacob Steinhardt.
\newblock Progress measures for grokking via mechanistic interpretability.
\newblock In \emph{The Eleventh International Conference on Learning Representations}, 2023.
\newblock URL \url{https://openreview.net/forum?id=9XFSbDPmdW}.

\bibitem[Newman et~al.(2021)Newman, Ang, Gong, and Hewitt]{newman-etal-2021-refining}
Benjamin Newman, Kai-Siang Ang, Julia Gong, and John Hewitt.
\newblock Refining targeted syntactic evaluation of language models.
\newblock In Kristina Toutanova, Anna Rumshisky, Luke Zettlemoyer, Dilek Hakkani-Tur, Iz~Beltagy, Steven Bethard, Ryan Cotterell, Tanmoy Chakraborty, and Yichao Zhou, editors, \emph{Proceedings of the 2021 Conference of the North American Chapter of the Association for Computational Linguistics: Human Language Technologies}, pages 3710--3723, Online, June 2021. Association for Computational Linguistics.
\newblock \doi{10.18653/v1/2021.naacl-main.290}.
\newblock URL \url{https://aclanthology.org/2021.naacl-main.290}.

\bibitem[Olah et~al.(2020)Olah, Cammarata, Schubert, Goh, Petrov, and Carter]{olah2020zoom}
Chris Olah, Nick Cammarata, Ludwig Schubert, Gabriel Goh, Michael Petrov, and Shan Carter.
\newblock Zoom in: An introduction to circuits.
\newblock \emph{Distill}, 2020.
\newblock \doi{10.23915/distill.00024.001}.
\newblock https://distill.pub/2020/circuits/zoom-in.

\bibitem[Olsson et~al.(2022)Olsson, Elhage, Nanda, Joseph, DasSarma, Henighan, Mann, Askell, Bai, Chen, Conerly, Drain, Ganguli, Hatfield-Dodds, Hernandez, Johnston, Jones, Kernion, Lovitt, Ndousse, Amodei, Brown, Clark, Kaplan, McCandlish, and Olah]{olsson2022context}
Catherine Olsson, Nelson Elhage, Neel Nanda, Nicholas Joseph, Nova DasSarma, Tom Henighan, Ben Mann, Amanda Askell, Yuntao Bai, Anna Chen, Tom Conerly, Dawn Drain, Deep Ganguli, Zac Hatfield-Dodds, Danny Hernandez, Scott Johnston, Andy Jones, Jackson Kernion, Liane Lovitt, Kamal Ndousse, Dario Amodei, Tom Brown, Jack Clark, Jared Kaplan, Sam McCandlish, and Chris Olah.
\newblock In-context learning and induction heads.
\newblock \emph{Transformer Circuits Thread}, 2022.
\newblock https://transformer-circuits.pub/2022/in-context-learning-and-induction-heads/index.html.

\bibitem[OpenAI et~al.(2024)OpenAI, Achiam, Adler, Agarwal, Ahmad, Akkaya, Aleman, Almeida, Altenschmidt, Altman, Anadkat, Avila, Babuschkin, Balaji, Balcom, Baltescu, Bao, Bavarian, Belgum, Bello, Berdine, Bernadett-Shapiro, Berner, Bogdonoff, Boiko, Boyd, Brakman, Brockman, Brooks, Brundage, Button, Cai, Campbell, Cann, Carey, Carlson, Carmichael, Chan, Chang, Chantzis, Chen, Chen, Chen, Chen, Chen, Chess, Cho, Chu, Chung, Cummings, Currier, Dai, Decareaux, Degry, Deutsch, Deville, Dhar, Dohan, Dowling, Dunning, Ecoffet, Eleti, Eloundou, Farhi, Fedus, Felix, Fishman, Forte, Fulford, Gao, Georges, Gibson, Goel, Gogineni, Goh, Gontijo-Lopes, Gordon, Grafstein, Gray, Greene, Gross, Gu, Guo, Hallacy, Han, Harris, He, Heaton, Heidecke, Hesse, Hickey, Hickey, Hoeschele, Houghton, Hsu, Hu, Hu, Huizinga, Jain, Jain, Jang, Jiang, Jiang, Jin, Jin, Jomoto, Jonn, Jun, Kaftan, Łukasz Kaiser, Kamali, Kanitscheider, Keskar, Khan, Kilpatrick, Kim, Kim, Kim, Kirchner, Kiros, Knight, Kokotajlo, Łukasz Kondraciuk, Kondrich,
  Konstantinidis, Kosic, Krueger, Kuo, Lampe, Lan, Lee, Leike, Leung, Levy, Li, Lim, Lin, Lin, Litwin, Lopez, Lowe, Lue, Makanju, Malfacini, Manning, Markov, Markovski, Martin, Mayer, Mayne, McGrew, McKinney, McLeavey, McMillan, McNeil, Medina, Mehta, Menick, Metz, Mishchenko, Mishkin, Monaco, Morikawa, Mossing, Mu, Murati, Murk, Mély, Nair, Nakano, Nayak, Neelakantan, Ngo, Noh, Ouyang, O'Keefe, Pachocki, Paino, Palermo, Pantuliano, Parascandolo, Parish, Parparita, Passos, Pavlov, Peng, Perelman, de~Avila Belbute~Peres, Petrov, de~Oliveira~Pinto, Michael, Pokorny, Pokrass, Pong, Powell, Power, Power, Proehl, Puri, Radford, Rae, Ramesh, Raymond, Real, Rimbach, Ross, Rotsted, Roussez, Ryder, Saltarelli, Sanders, Santurkar, Sastry, Schmidt, Schnurr, Schulman, Selsam, Sheppard, Sherbakov, Shieh, Shoker, Shyam, Sidor, Sigler, Simens, Sitkin, Slama, Sohl, Sokolowsky, Song, Staudacher, Such, Summers, Sutskever, Tang, Tezak, Thompson, Tillet, Tootoonchian, Tseng, Tuggle, Turley, Tworek, Uribe, Vallone, Vijayvergiya,
  Voss, Wainwright, Wang, Wang, Wang, Ward, Wei, Weinmann, Welihinda, Welinder, Weng, Weng, Wiethoff, Willner, Winter, Wolrich, Wong, Workman, Wu, Wu, Wu, Xiao, Xu, Yoo, Yu, Yuan, Zaremba, Zellers, Zhang, Zhang, Zhao, Zheng, Zhuang, Zhuk, and Zoph]{openai2024gpt4}
OpenAI, Josh Achiam, Steven Adler, Sandhini Agarwal, Lama Ahmad, Ilge Akkaya, Florencia~Leoni Aleman, Diogo Almeida, Janko Altenschmidt, Sam Altman, Shyamal Anadkat, Red Avila, Igor Babuschkin, Suchir Balaji, Valerie Balcom, Paul Baltescu, Haiming Bao, Mohammad Bavarian, Jeff Belgum, Irwan Bello, Jake Berdine, Gabriel Bernadett-Shapiro, Christopher Berner, Lenny Bogdonoff, Oleg Boiko, Madelaine Boyd, Anna-Luisa Brakman, Greg Brockman, Tim Brooks, Miles Brundage, Kevin Button, Trevor Cai, Rosie Campbell, Andrew Cann, Brittany Carey, Chelsea Carlson, Rory Carmichael, Brooke Chan, Che Chang, Fotis Chantzis, Derek Chen, Sully Chen, Ruby Chen, Jason Chen, Mark Chen, Ben Chess, Chester Cho, Casey Chu, Hyung~Won Chung, Dave Cummings, Jeremiah Currier, Yunxing Dai, Cory Decareaux, Thomas Degry, Noah Deutsch, Damien Deville, Arka Dhar, David Dohan, Steve Dowling, Sheila Dunning, Adrien Ecoffet, Atty Eleti, Tyna Eloundou, David Farhi, Liam Fedus, Niko Felix, Simón~Posada Fishman, Juston Forte, Isabella Fulford, Leo
  Gao, Elie Georges, Christian Gibson, Vik Goel, Tarun Gogineni, Gabriel Goh, Rapha Gontijo-Lopes, Jonathan Gordon, Morgan Grafstein, Scott Gray, Ryan Greene, Joshua Gross, Shixiang~Shane Gu, Yufei Guo, Chris Hallacy, Jesse Han, Jeff Harris, Yuchen He, Mike Heaton, Johannes Heidecke, Chris Hesse, Alan Hickey, Wade Hickey, Peter Hoeschele, Brandon Houghton, Kenny Hsu, Shengli Hu, Xin Hu, Joost Huizinga, Shantanu Jain, Shawn Jain, Joanne Jang, Angela Jiang, Roger Jiang, Haozhun Jin, Denny Jin, Shino Jomoto, Billie Jonn, Heewoo Jun, Tomer Kaftan, Łukasz Kaiser, Ali Kamali, Ingmar Kanitscheider, Nitish~Shirish Keskar, Tabarak Khan, Logan Kilpatrick, Jong~Wook Kim, Christina Kim, Yongjik Kim, Jan~Hendrik Kirchner, Jamie Kiros, Matt Knight, Daniel Kokotajlo, Łukasz Kondraciuk, Andrew Kondrich, Aris Konstantinidis, Kyle Kosic, Gretchen Krueger, Vishal Kuo, Michael Lampe, Ikai Lan, Teddy Lee, Jan Leike, Jade Leung, Daniel Levy, Chak~Ming Li, Rachel Lim, Molly Lin, Stephanie Lin, Mateusz Litwin, Theresa Lopez, Ryan
  Lowe, Patricia Lue, Anna Makanju, Kim Malfacini, Sam Manning, Todor Markov, Yaniv Markovski, Bianca Martin, Katie Mayer, Andrew Mayne, Bob McGrew, Scott~Mayer McKinney, Christine McLeavey, Paul McMillan, Jake McNeil, David Medina, Aalok Mehta, Jacob Menick, Luke Metz, Andrey Mishchenko, Pamela Mishkin, Vinnie Monaco, Evan Morikawa, Daniel Mossing, Tong Mu, Mira Murati, Oleg Murk, David Mély, Ashvin Nair, Reiichiro Nakano, Rajeev Nayak, Arvind Neelakantan, Richard Ngo, Hyeonwoo Noh, Long Ouyang, Cullen O'Keefe, Jakub Pachocki, Alex Paino, Joe Palermo, Ashley Pantuliano, Giambattista Parascandolo, Joel Parish, Emy Parparita, Alex Passos, Mikhail Pavlov, Andrew Peng, Adam Perelman, Filipe de~Avila Belbute~Peres, Michael Petrov, Henrique~Ponde de~Oliveira~Pinto, Michael, Pokorny, Michelle Pokrass, Vitchyr~H. Pong, Tolly Powell, Alethea Power, Boris Power, Elizabeth Proehl, Raul Puri, Alec Radford, Jack Rae, Aditya Ramesh, Cameron Raymond, Francis Real, Kendra Rimbach, Carl Ross, Bob Rotsted, Henri Roussez,
  Nick Ryder, Mario Saltarelli, Ted Sanders, Shibani Santurkar, Girish Sastry, Heather Schmidt, David Schnurr, John Schulman, Daniel Selsam, Kyla Sheppard, Toki Sherbakov, Jessica Shieh, Sarah Shoker, Pranav Shyam, Szymon Sidor, Eric Sigler, Maddie Simens, Jordan Sitkin, Katarina Slama, Ian Sohl, Benjamin Sokolowsky, Yang Song, Natalie Staudacher, Felipe~Petroski Such, Natalie Summers, Ilya Sutskever, Jie Tang, Nikolas Tezak, Madeleine~B. Thompson, Phil Tillet, Amin Tootoonchian, Elizabeth Tseng, Preston Tuggle, Nick Turley, Jerry Tworek, Juan Felipe~Cerón Uribe, Andrea Vallone, Arun Vijayvergiya, Chelsea Voss, Carroll Wainwright, Justin~Jay Wang, Alvin Wang, Ben Wang, Jonathan Ward, Jason Wei, CJ~Weinmann, Akila Welihinda, Peter Welinder, Jiayi Weng, Lilian Weng, Matt Wiethoff, Dave Willner, Clemens Winter, Samuel Wolrich, Hannah Wong, Lauren Workman, Sherwin Wu, Jeff Wu, Michael Wu, Kai Xiao, Tao Xu, Sarah Yoo, Kevin Yu, Qiming Yuan, Wojciech Zaremba, Rowan Zellers, Chong Zhang, Marvin Zhang, Shengjia
  Zhao, Tianhao Zheng, Juntang Zhuang, William Zhuk, and Barret Zoph.
\newblock Gpt-4 technical report, 2024.

\bibitem[Prakash et~al.(2024)Prakash, Shaham, Haklay, Belinkov, and Bau]{prakash2024finetuning}
Nikhil Prakash, Tamar~Rott Shaham, Tal Haklay, Yonatan Belinkov, and David Bau.
\newblock Fine-tuning enhances existing mechanisms: A case study on entity tracking.
\newblock In \emph{The Twelfth International Conference on Learning Representations}, 2024.
\newblock URL \url{https://openreview.net/forum?id=8sKcAWOf2D}.

\bibitem[Rae et~al.(2022)Rae, Borgeaud, Cai, Millican, Hoffmann, Song, Aslanides, Henderson, Ring, Young, Rutherford, Hennigan, Menick, Cassirer, Powell, van~den Driessche, Hendricks, Rauh, Huang, Glaese, Welbl, Dathathri, Huang, Uesato, Mellor, Higgins, Creswell, McAleese, Wu, Elsen, Jayakumar, Buchatskaya, Budden, Sutherland, Simonyan, Paganini, Sifre, Martens, Li, Kuncoro, Nematzadeh, Gribovskaya, Donato, Lazaridou, Mensch, Lespiau, Tsimpoukelli, Grigorev, Fritz, Sottiaux, Pajarskas, Pohlen, Gong, Toyama, de~Masson~d'Autume, Li, Terzi, Mikulik, Babuschkin, Clark, de~Las~Casas, Guy, Jones, Bradbury, Johnson, Hechtman, Weidinger, Gabriel, Isaac, Lockhart, Osindero, Rimell, Dyer, Vinyals, Ayoub, Stanway, Bennett, Hassabis, Kavukcuoglu, and Irving]{rae2022scaling}
Jack~W. Rae, Sebastian Borgeaud, Trevor Cai, Katie Millican, Jordan Hoffmann, Francis Song, John Aslanides, Sarah Henderson, Roman Ring, Susannah Young, Eliza Rutherford, Tom Hennigan, Jacob Menick, Albin Cassirer, Richard Powell, George van~den Driessche, Lisa~Anne Hendricks, Maribeth Rauh, Po-Sen Huang, Amelia Glaese, Johannes Welbl, Sumanth Dathathri, Saffron Huang, Jonathan Uesato, John Mellor, Irina Higgins, Antonia Creswell, Nat McAleese, Amy Wu, Erich Elsen, Siddhant Jayakumar, Elena Buchatskaya, David Budden, Esme Sutherland, Karen Simonyan, Michela Paganini, Laurent Sifre, Lena Martens, Xiang~Lorraine Li, Adhiguna Kuncoro, Aida Nematzadeh, Elena Gribovskaya, Domenic Donato, Angeliki Lazaridou, Arthur Mensch, Jean-Baptiste Lespiau, Maria Tsimpoukelli, Nikolai Grigorev, Doug Fritz, Thibault Sottiaux, Mantas Pajarskas, Toby Pohlen, Zhitao Gong, Daniel Toyama, Cyprien de~Masson~d'Autume, Yujia Li, Tayfun Terzi, Vladimir Mikulik, Igor Babuschkin, Aidan Clark, Diego de~Las~Casas, Aurelia Guy, Chris Jones,
  James Bradbury, Matthew Johnson, Blake Hechtman, Laura Weidinger, Iason Gabriel, William Isaac, Ed~Lockhart, Simon Osindero, Laura Rimell, Chris Dyer, Oriol Vinyals, Kareem Ayoub, Jeff Stanway, Lorrayne Bennett, Demis Hassabis, Koray Kavukcuoglu, and Geoffrey Irving.
\newblock Scaling language models: Methods, analysis \& insights from training gopher, 2022.

\bibitem[Rushing and Nanda(2024)]{rushing2024explorations}
Cody Rushing and Neel Nanda.
\newblock Explorations of self-repair in language models, 2024.

\bibitem[Sajjad et~al.(2022)Sajjad, Durrani, and Dalvi]{sajjad2022neuronlevel}
Hassan Sajjad, Nadir Durrani, and Fahim Dalvi.
\newblock Neuron-level interpretation of deep nlp models: A survey, 2022.

\bibitem[Sakaguchi et~al.(2019)Sakaguchi, Bras, Bhagavatula, and Choi]{winogrande}
Keisuke Sakaguchi, Ronan~Le Bras, Chandra Bhagavatula, and Yejin Choi.
\newblock Winogrande: An adversarial winograd schema challenge at scale, 2019.

\bibitem[Sellam et~al.(2022)Sellam, Yadlowsky, Tenney, Wei, Saphra, D'Amour, Linzen, Bastings, Turc, Eisenstein, Das, and Pavlick]{sellam2022the}
Thibault Sellam, Steve Yadlowsky, Ian Tenney, Jason Wei, Naomi Saphra, Alexander D'Amour, Tal Linzen, Jasmijn Bastings, Iulia~Raluca Turc, Jacob Eisenstein, Dipanjan Das, and Ellie Pavlick.
\newblock The multi{BERT}s: {BERT} reproductions for robustness analysis.
\newblock In \emph{International Conference on Learning Representations}, 2022.
\newblock URL \url{https://openreview.net/forum?id=K0E_F0gFDgA}.

\bibitem[Shrikumar et~al.(2017)Shrikumar, Greenside, and Kundaje]{pmlr-v70-shrikumar17a}
Avanti Shrikumar, Peyton Greenside, and Anshul Kundaje.
\newblock Learning important features through propagating activation differences.
\newblock In Doina Precup and Yee~Whye Teh, editors, \emph{Proceedings of the 34th International Conference on Machine Learning}, volume~70 of \emph{Proceedings of Machine Learning Research}, pages 3145--3153. PMLR, 06--11 Aug 2017.
\newblock URL \url{https://proceedings.mlr.press/v70/shrikumar17a.html}.

\bibitem[Singh et~al.(2024)Singh, Moskovitz, Hill, Chan, and Saxe]{singh2024needs}
Aaditya~K. Singh, Ted Moskovitz, Felix Hill, Stephanie C.~Y. Chan, and Andrew~M. Saxe.
\newblock What needs to go right for an induction head? a mechanistic study of in-context learning circuits and their formation, 2024.

\bibitem[Sundararajan et~al.(2017)Sundararajan, Taly, and Yan]{Sundararajan2017AxiomaticAF}
Mukund Sundararajan, Ankur Taly, and Qiqi Yan.
\newblock Axiomatic attribution for deep networks.
\newblock In \emph{International Conference on Machine Learning}, 2017.
\newblock URL \url{https://proceedings.mlr.press/v70/sundararajan17a/sundararajan17a.pdf}.

\bibitem[Syed et~al.(2023)Syed, Rager, and Conmy]{syed2023attribution}
Aaquib Syed, Can Rager, and Arthur Conmy.
\newblock Attribution patching outperforms automated circuit discovery, 2023.

\bibitem[Team et~al.(2024)Team, Anil, Borgeaud, Alayrac, Yu, Soricut, Schalkwyk, Dai, Hauth, Millican, Silver, Johnson, Antonoglou, Schrittwieser, Glaese, Chen, Pitler, Lillicrap, Lazaridou, Firat, Molloy, Isard, Barham, Hennigan, Lee, Viola, Reynolds, Xu, Doherty, Collins, Meyer, Rutherford, Moreira, Ayoub, Goel, Krawczyk, Du, Chi, Cheng, Ni, Shah, Kane, Chan, Faruqui, Severyn, Lin, Li, Cheng, Ittycheriah, Mahdieh, Chen, Sun, Tran, Bagri, Lakshminarayanan, Liu, Orban, Güra, Zhou, Song, Boffy, Ganapathy, Zheng, Choe, Ágoston Weisz, Zhu, Lu, Gopal, Kahn, Kula, Pitman, Shah, Taropa, Merey, Baeuml, Chen, Shafey, Zhang, Sercinoglu, Tucker, Piqueras, Krikun, Barr, Savinov, Danihelka, Roelofs, White, Andreassen, von Glehn, Yagati, Kazemi, Gonzalez, Khalman, Sygnowski, Frechette, Smith, Culp, Proleev, Luan, Chen, Lottes, Schucher, Lebron, Rrustemi, Clay, Crone, Kocisky, Zhao, Perz, Yu, Howard, Bloniarz, Rae, Lu, Sifre, Maggioni, Alcober, Garrette, Barnes, Thakoor, Austin, Barth-Maron, Wong, Joshi, Chaabouni,
  Fatiha, Ahuja, Tomar, Senter, Chadwick, Kornakov, Attaluri, Iturrate, Liu, Li, Cogan, Chen, Jia, Gu, Zhang, Grimstad, Hartman, Garcia, Pillai, Devlin, Laskin, de~Las~Casas, Valter, Tao, Blanco, Badia, Reitter, Chen, Brennan, Rivera, Brin, Iqbal, Surita, Labanowski, Rao, Winkler, Parisotto, Gu, Olszewska, Addanki, Miech, Louis, Teplyashin, Brown, Catt, Balaguer, Xiang, Wang, Ashwood, Briukhov, Webson, Ganapathy, Sanghavi, Kannan, Chang, Stjerngren, Djolonga, Sun, Bapna, Aitchison, Pejman, Michalewski, Yu, Wang, Love, Ahn, Bloxwich, Han, Humphreys, Sellam, Bradbury, Godbole, Samangooei, Damoc, Kaskasoli, Arnold, Vasudevan, Agrawal, Riesa, Lepikhin, Tanburn, Srinivasan, Lim, Hodkinson, Shyam, Ferret, Hand, Garg, Paine, Li, Li, Giang, Neitz, Abbas, York, Reid, Cole, Chowdhery, Das, Rogozińska, Nikolaev, Sprechmann, Nado, Zilka, Prost, He, Monteiro, Mishra, Welty, Newlan, Jia, Allamanis, Hu, de~Liedekerke, Gilmer, Saroufim, Rijhwani, Hou, Shrivastava, Baddepudi, Goldin, Ozturel, Cassirer, Xu, Sohn, Sachan,
  Amplayo, Swanson, Petrova, Narayan, Guez, Brahma, Landon, Patel, Zhao, Villela, Wang, Jia, Rahtz, Giménez, Yeung, Keeling, Georgiev, Mincu, Wu, Haykal, Saputro, Vodrahalli, Qin, Cankara, Sharma, Fernando, Hawkins, Neyshabur, Kim, Hutter, Agrawal, Castro-Ros, van~den Driessche, Wang, Yang, yiin Chang, Komarek, McIlroy, Lučić, Zhang, Farhan, Sharman, Natsev, Michel, Bansal, Qiao, Cao, Shakeri, Butterfield, Chung, Rubenstein, Agrawal, Mensch, Soparkar, Lenc, Chung, Pope, Maggiore, Kay, Jhakra, Wang, Maynez, Phuong, Tobin, Tacchetti, Trebacz, Robinson, Katariya, Riedel, Bailey, Xiao, Ghelani, Aroyo, Slone, Houlsby, Xiong, Yang, Gribovskaya, Adler, Wirth, Lee, Li, Kagohara, Pavagadhi, Bridgers, Bortsova, Ghemawat, Ahmed, Liu, Powell, Bolina, Iinuma, Zablotskaia, Besley, Chung, Dozat, Comanescu, Si, Greer, Su, Polacek, Kaufman, Tokumine, Hu, Buchatskaya, Miao, Elhawaty, Siddhant, Tomasev, Xing, Greer, Miller, Ashraf, Roy, Zhang, Ma, Filos, Besta, Blevins, Klimenko, Yeh, Changpinyo, Mu, Chang, Pajarskas, Muir,
  Cohen, Lan, Haridasan, Marathe, Hansen, Douglas, Samuel, Wang, Austin, Lan, Jiang, Chiu, Lorenzo, Sjösund, Cevey, Gleicher, Avrahami, Boral, Srinivasan, Selo, May, Aisopos, Hussenot, Soares, Baumli, Chang, Recasens, Caine, Pritzel, Pavetic, Pardo, Gergely, Frye, Ramasesh, Horgan, Badola, Kassner, Roy, Dyer, Campos, Tomala, Tang, Badawy, White, Mustafa, Lang, Jindal, Vikram, Gong, Caelles, Hemsley, Thornton, Feng, Stokowiec, Zheng, Thacker, Çağlar Ünlü, Zhang, Saleh, Svensson, Bileschi, Patil, Anand, Ring, Tsihlas, Vezer, Selvi, Shevlane, Rodriguez, Kwiatkowski, Daruki, Rong, Dafoe, FitzGerald, Gu-Lemberg, Khan, Hendricks, Pellat, Feinberg, Cobon-Kerr, Sainath, Rauh, Hashemi, Ives, Hasson, Noland, Cao, Byrd, Hou, Wang, Sottiaux, Paganini, Lespiau, Moufarek, Hassan, Shivakumar, van Amersfoort, Mandhane, Joshi, Goyal, Tung, Brock, Sheahan, Misra, Li, Rakićević, Dehghani, Liu, Mittal, Oh, Noury, Sezener, Huot, Lamm, Cao, Chen, Mudgal, Stella, Brooks, Vasudevan, Liu, Chain, Melinkeri, Cohen, Wang,
  Seymore, Zubkov, Goel, Yue, Krishnakumaran, Albert, Hurley, Sano, Mohananey, Joughin, Filonov, Kępa, Eldawy, Lim, Rishi, Badiezadegan, Bos, Chang, Jain, Padmanabhan, Puttagunta, Krishna, Baker, Kalb, Bedapudi, Kurzrok, Lei, Yu, Litvin, Zhou, Wu, Sobell, Siciliano, Papir, Neale, Bragagnolo, Toor, Chen, Anklin, Wang, Feng, Gholami, Ling, Liu, Walter, Moghaddam, Kishore, Adamek, Mercado, Mallinson, Wandekar, Cagle, Ofek, Garrido, Lombriser, Mukha, Sun, Mohammad, Matak, Qian, Peswani, Janus, Yuan, Schelin, David, Garg, He, Duzhyi, Älgmyr, Lottaz, Li, Yadav, Xu, Chinien, Shivanna, Chuklin, Li, Spadine, Wolfe, Mohamed, Das, Dai, He, von Dincklage, Upadhyay, Maurya, Chi, Krause, Salama, Rabinovitch, M, Selvan, Dektiarev, Ghiasi, Guven, Gupta, Liu, Sharma, Shtacher, Paul, Akerlund, Aubet, Huang, Zhu, Zhu, Teixeira, Fritze, Bertolini, Marinescu, Bölle, Paulus, Gupta, Latkar, Chang, Sanders, Wilson, Wu, Tan, Thiet, Doshi, Lall, Mishra, Chen, Luong, Benjamin, Lee, Andrejczuk, Rabiej, Ranjan, Styrc, Yin, Simon,
  Harriott, Bansal, Robsky, Bacon, Greene, Mirylenka, Zhou, Sarvana, Goyal, Andermatt, Siegler, Horn, Israel, Pongetti, Chen, Selvatici, Silva, Wang, Tolins, Guu, Yogev, Cai, Agostini, Shah, Nguyen, Donnaile, Pereira, Friso, Stambler, Kurzrok, Kuang, Romanikhin, Geller, Yan, Jang, Lee, Fica, Malmi, Tan, Banica, Balle, Pham, Huang, Avram, Shi, Singh, Hidey, Ahuja, Saxena, Dooley, Potharaju, O'Neill, Gokulchandran, Foley, Zhao, Dusenberry, Liu, Mehta, Kotikalapudi, Safranek-Shrader, Goodman, Kessinger, Globen, Kolhar, Gorgolewski, Ibrahim, Song, Eichenbaum, Brovelli, Potluri, Lahoti, Baetu, Ghorbani, Chen, Crawford, Pal, Sridhar, Gurita, Mujika, Petrovski, Cedoz, Li, Chen, Santo, Goyal, Punjabi, Kappaganthu, Kwak, LV, Velury, Choudhury, Hall, Shah, Figueira, Thomas, Lu, Zhou, Kumar, Jurdi, Chikkerur, Ma, Yu, Kwak, Ähdel, Rajayogam, Choma, Liu, Barua, Ji, Park, Hellendoorn, Bailey, Bilal, Zhou, Khatir, Sutton, Rzadkowski, Macintosh, Shagin, Medina, Liang, Zhou, Shah, Bi, Dankovics, Banga, Lehmann, Bredesen,
  Lin, Hoffmann, Lai, Chung, Yang, Balani, Bražinskas, Sozanschi, Hayes, Alcalde, Makarov, Chen, Stella, Snijders, Mandl, Kärrman, Nowak, Wu, Dyck, Vaidyanathan, R, Mallet, Rudominer, Johnston, Mittal, Udathu, Christensen, Verma, Irving, Santucci, Elsayed, Davoodi, Georgiev, Tenney, Hua, Cideron, Leurent, Alnahlawi, Georgescu, Wei, Zheng, Scandinaro, Jiang, Snoek, Sundararajan, Wang, Ontiveros, Karo, Cole, Rajashekhar, Tumeh, Ben-David, Jain, Uesato, Datta, Bunyan, Wu, Zhang, Stanczyk, Zhang, Steiner, Naskar, Azzam, Johnson, Paszke, Chiu, Elias, Mohiuddin, Muhammad, Miao, Lee, Vieillard, Park, Zhang, Stanway, Garmon, Karmarkar, Dong, Lee, Kumar, Zhou, Evens, Isaac, Irving, Loper, Fink, Arkatkar, Chen, Shafran, Petrychenko, Chen, Jia, Levskaya, Zhu, Grabowski, Mao, Magni, Yao, Snaider, Casagrande, Palmer, Suganthan, Castaño, Giannoumis, Kim, Rybiński, Sreevatsa, Prendki, Soergel, Goedeckemeyer, Gierke, Jafari, Gaba, Wiesner, Wright, Wei, Vashisht, Kulizhskaya, Hoover, Le, Li, Iwuanyanwu, Liu, Ramirez,
  Khorlin, Cui, LIN, Wu, Aguilar, Pallo, Chakladar, Perng, Abellan, Zhang, Dasgupta, Kushman, Penchev, Repina, Wu, van~der Weide, Ponnapalli, Kaplan, Simsa, Li, Dousse, Yang, Piper, Ie, Pasumarthi, Lintz, Vijayakumar, Andor, Valenzuela, Lui, Paduraru, Peng, Lee, Zhang, Greene, Nguyen, Kurylowicz, Hardin, Dixon, Janzer, Choo, Feng, Zhang, Singhal, Du, McKinnon, Antropova, Bolukbasi, Keller, Reid, Finchelstein, Raad, Crocker, Hawkins, Dadashi, Gaffney, Franko, Bulanova, Leblond, Chung, Askham, Cobo, Xu, Fischer, Xu, Sorokin, Alberti, Lin, Evans, Dimitriev, Forbes, Banarse, Tung, Omernick, Bishop, Sterneck, Jain, Xia, Amid, Piccinno, Wang, Banzal, Mankowitz, Polozov, Krakovna, Brown, Bateni, Duan, Firoiu, Thotakuri, Natan, Geist, tan Girgin, Li, Ye, Roval, Tojo, Kwong, Lee-Thorp, Yew, Sinopalnikov, Ramos, Mellor, Sharma, Wu, Miller, Sonnerat, Vnukov, Greig, Beattie, Caveness, Bai, Eisenschlos, Korchemniy, Tsai, Jasarevic, Kong, Dao, Zheng, Liu, Yang, Zhu, Teh, Sanmiya, Gladchenko, Trdin, Toyama, Rosen, Tavakkol,
  Xue, Elkind, Woodman, Carpenter, Papamakarios, Kemp, Kafle, Grunina, Sinha, Talbert, Wu, Owusu-Afriyie, Du, Thornton, Pont-Tuset, Narayana, Li, Fatehi, Wieting, Ajmeri, Uria, Ko, Knight, Héliou, Niu, Gu, Pang, Li, Levine, Stolovich, Santamaria-Fernandez, Goenka, Yustalim, Strudel, Elqursh, Deck, Lee, Li, Levin, Hoffmann, Holtmann-Rice, Bachem, Arora, Koh, Yeganeh, Põder, Tariq, Sun, Ionita, Seyedhosseini, Tafti, Liu, Gulati, Liu, Ye, Chrzaszcz, Wang, Sethi, Li, Brown, Singh, Fan, Parisi, Stanton, Koverkathu, Choquette-Choo, Li, Lu, Ittycheriah, Shroff, Varadarajan, Bahargam, Willoughby, Gaddy, Desjardins, Cornero, Robenek, Mittal, Albrecht, Shenoy, Moiseev, Jacobsson, Ghaffarkhah, Rivière, Walton, Crepy, Parrish, Zhou, Farabet, Radebaugh, Srinivasan, van~der Salm, Fidjeland, Scellato, Latorre-Chimoto, Klimczak-Plucińska, Bridson, de~Cesare, Hudson, Mendolicchio, Walker, Morris, Mauger, Guseynov, Reid, Odoom, Loher, Cotruta, Yenugula, Grewe, Petrushkina, Duerig, Sanchez, Yadlowsky, Shen, Globerson, Webb,
  Dua, Li, Bhupatiraju, Hurt, Qureshi, Agarwal, Shani, Eyal, Khare, Belle, Wang, Tekur, Kale, Wei, Sang, Saeta, Liechty, Sun, Zhao, Lee, Nayak, Fritz, Vuyyuru, Aslanides, Vyas, Wicke, Ma, Eltyshev, Martin, Cate, Manyika, Amiri, Kim, Xiong, Kang, Luisier, Tripuraneni, Madras, Guo, Waters, Wang, Ainslie, Baldridge, Zhang, Pruthi, Bauer, Yang, Mansour, Gelman, Xu, Polovets, Liu, Cai, Chen, Sheng, Xue, Ozair, Angermueller, Li, Sinha, Wang, Wiesinger, Koukoumidis, Tian, Iyer, Gurumurthy, Goldenson, Shah, Blake, Yu, Urbanowicz, Palomaki, Fernando, Durden, Mehta, Momchev, Rahimtoroghi, Georgaki, Raul, Ruder, Redshaw, Lee, Zhou, Jalan, Li, Hechtman, Schuh, Nasr, Milan, Mikulik, Franco, Green, Nguyen, Kelley, Mahendru, Hu, Howland, Vargas, Hui, Bansal, Rao, Ghiya, Wang, Ye, Sarr, Preston, Elish, Li, Kaku, Gupta, Pasupat, Juan, Someswar, M., Chen, Amini, Fabrikant, Chu, Dong, Muthal, Buthpitiya, Jauhari, Hua, Khandelwal, Hitron, Ren, Rinaldi, Drath, Dabush, Jiang, Godhia, Sachs, Chen, Fan, Taitelbaum, Noga, Dai, Wang,
  Liang, Hamer, Ferng, Elkind, Atias, Lee, Listík, Carlen, van~de Kerkhof, Pikus, Zaher, Müller, Zykova, Stefanec, Gatsko, Hirnschall, Sethi, Xu, Ahuja, Tsai, Stefanoiu, Feng, Dhandhania, Katyal, Gupta, Parulekar, Pitta, Zhao, Bhatia, Bhavnani, Alhadlaq, Li, Danenberg, Tu, Pine, Filippova, Ghosh, Limonchik, Urala, Lanka, Clive, Sun, Li, Wu, Hongtongsak, Li, Thakkar, Omarov, Majmundar, Alverson, Kucharski, Patel, Jain, Zabelin, Pelagatti, Kohli, Kumar, Kim, Sankar, Shah, Ramachandruni, Zeng, Bariach, Weidinger, Subramanya, Hsiao, Hassabis, Kavukcuoglu, Sadovsky, Le, Strohman, Wu, Petrov, Dean, and Vinyals]{geminiteam2024gemini}
Gemini Team, Rohan Anil, Sebastian Borgeaud, Jean-Baptiste Alayrac, Jiahui Yu, Radu Soricut, Johan Schalkwyk, Andrew~M. Dai, Anja Hauth, Katie Millican, David Silver, Melvin Johnson, Ioannis Antonoglou, Julian Schrittwieser, Amelia Glaese, Jilin Chen, Emily Pitler, Timothy Lillicrap, Angeliki Lazaridou, Orhan Firat, James Molloy, Michael Isard, Paul~R. Barham, Tom Hennigan, Benjamin Lee, Fabio Viola, Malcolm Reynolds, Yuanzhong Xu, Ryan Doherty, Eli Collins, Clemens Meyer, Eliza Rutherford, Erica Moreira, Kareem Ayoub, Megha Goel, Jack Krawczyk, Cosmo Du, Ed~Chi, Heng-Tze Cheng, Eric Ni, Purvi Shah, Patrick Kane, Betty Chan, Manaal Faruqui, Aliaksei Severyn, Hanzhao Lin, YaGuang Li, Yong Cheng, Abe Ittycheriah, Mahdis Mahdieh, Mia Chen, Pei Sun, Dustin Tran, Sumit Bagri, Balaji Lakshminarayanan, Jeremiah Liu, Andras Orban, Fabian Güra, Hao Zhou, Xinying Song, Aurelien Boffy, Harish Ganapathy, Steven Zheng, HyunJeong Choe, Ágoston Weisz, Tao Zhu, Yifeng Lu, Siddharth Gopal, Jarrod Kahn, Maciej Kula, Jeff
  Pitman, Rushin Shah, Emanuel Taropa, Majd~Al Merey, Martin Baeuml, Zhifeng Chen, Laurent~El Shafey, Yujing Zhang, Olcan Sercinoglu, George Tucker, Enrique Piqueras, Maxim Krikun, Iain Barr, Nikolay Savinov, Ivo Danihelka, Becca Roelofs, Anaïs White, Anders Andreassen, Tamara von Glehn, Lakshman Yagati, Mehran Kazemi, Lucas Gonzalez, Misha Khalman, Jakub Sygnowski, Alexandre Frechette, Charlotte Smith, Laura Culp, Lev Proleev, Yi~Luan, Xi~Chen, James Lottes, Nathan Schucher, Federico Lebron, Alban Rrustemi, Natalie Clay, Phil Crone, Tomas Kocisky, Jeffrey Zhao, Bartek Perz, Dian Yu, Heidi Howard, Adam Bloniarz, Jack~W. Rae, Han Lu, Laurent Sifre, Marcello Maggioni, Fred Alcober, Dan Garrette, Megan Barnes, Shantanu Thakoor, Jacob Austin, Gabriel Barth-Maron, William Wong, Rishabh Joshi, Rahma Chaabouni, Deeni Fatiha, Arun Ahuja, Gaurav~Singh Tomar, Evan Senter, Martin Chadwick, Ilya Kornakov, Nithya Attaluri, Iñaki Iturrate, Ruibo Liu, Yunxuan Li, Sarah Cogan, Jeremy Chen, Chao Jia, Chenjie Gu, Qiao Zhang,
  Jordan Grimstad, Ale~Jakse Hartman, Xavier Garcia, Thanumalayan~Sankaranarayana Pillai, Jacob Devlin, Michael Laskin, Diego de~Las~Casas, Dasha Valter, Connie Tao, Lorenzo Blanco, Adrià~Puigdomènech Badia, David Reitter, Mianna Chen, Jenny Brennan, Clara Rivera, Sergey Brin, Shariq Iqbal, Gabriela Surita, Jane Labanowski, Abhi Rao, Stephanie Winkler, Emilio Parisotto, Yiming Gu, Kate Olszewska, Ravi Addanki, Antoine Miech, Annie Louis, Denis Teplyashin, Geoff Brown, Elliot Catt, Jan Balaguer, Jackie Xiang, Pidong Wang, Zoe Ashwood, Anton Briukhov, Albert Webson, Sanjay Ganapathy, Smit Sanghavi, Ajay Kannan, Ming-Wei Chang, Axel Stjerngren, Josip Djolonga, Yuting Sun, Ankur Bapna, Matthew Aitchison, Pedram Pejman, Henryk Michalewski, Tianhe Yu, Cindy Wang, Juliette Love, Junwhan Ahn, Dawn Bloxwich, Kehang Han, Peter Humphreys, Thibault Sellam, James Bradbury, Varun Godbole, Sina Samangooei, Bogdan Damoc, Alex Kaskasoli, Sébastien M.~R. Arnold, Vijay Vasudevan, Shubham Agrawal, Jason Riesa, Dmitry
  Lepikhin, Richard Tanburn, Srivatsan Srinivasan, Hyeontaek Lim, Sarah Hodkinson, Pranav Shyam, Johan Ferret, Steven Hand, Ankush Garg, Tom~Le Paine, Jian Li, Yujia Li, Minh Giang, Alexander Neitz, Zaheer Abbas, Sarah York, Machel Reid, Elizabeth Cole, Aakanksha Chowdhery, Dipanjan Das, Dominika Rogozińska, Vitaliy Nikolaev, Pablo Sprechmann, Zachary Nado, Lukas Zilka, Flavien Prost, Luheng He, Marianne Monteiro, Gaurav Mishra, Chris Welty, Josh Newlan, Dawei Jia, Miltiadis Allamanis, Clara~Huiyi Hu, Raoul de~Liedekerke, Justin Gilmer, Carl Saroufim, Shruti Rijhwani, Shaobo Hou, Disha Shrivastava, Anirudh Baddepudi, Alex Goldin, Adnan Ozturel, Albin Cassirer, Yunhan Xu, Daniel Sohn, Devendra Sachan, Reinald~Kim Amplayo, Craig Swanson, Dessie Petrova, Shashi Narayan, Arthur Guez, Siddhartha Brahma, Jessica Landon, Miteyan Patel, Ruizhe Zhao, Kevin Villela, Luyu Wang, Wenhao Jia, Matthew Rahtz, Mai Giménez, Legg Yeung, James Keeling, Petko Georgiev, Diana Mincu, Boxi Wu, Salem Haykal, Rachel Saputro, Kiran
  Vodrahalli, James Qin, Zeynep Cankara, Abhanshu Sharma, Nick Fernando, Will Hawkins, Behnam Neyshabur, Solomon Kim, Adrian Hutter, Priyanka Agrawal, Alex Castro-Ros, George van~den Driessche, Tao Wang, Fan Yang, Shuo yiin Chang, Paul Komarek, Ross McIlroy, Mario Lučić, Guodong Zhang, Wael Farhan, Michael Sharman, Paul Natsev, Paul Michel, Yamini Bansal, Siyuan Qiao, Kris Cao, Siamak Shakeri, Christina Butterfield, Justin Chung, Paul~Kishan Rubenstein, Shivani Agrawal, Arthur Mensch, Kedar Soparkar, Karel Lenc, Timothy Chung, Aedan Pope, Loren Maggiore, Jackie Kay, Priya Jhakra, Shibo Wang, Joshua Maynez, Mary Phuong, Taylor Tobin, Andrea Tacchetti, Maja Trebacz, Kevin Robinson, Yash Katariya, Sebastian Riedel, Paige Bailey, Kefan Xiao, Nimesh Ghelani, Lora Aroyo, Ambrose Slone, Neil Houlsby, Xuehan Xiong, Zhen Yang, Elena Gribovskaya, Jonas Adler, Mateo Wirth, Lisa Lee, Music Li, Thais Kagohara, Jay Pavagadhi, Sophie Bridgers, Anna Bortsova, Sanjay Ghemawat, Zafarali Ahmed, Tianqi Liu, Richard Powell,
  Vijay Bolina, Mariko Iinuma, Polina Zablotskaia, James Besley, Da-Woon Chung, Timothy Dozat, Ramona Comanescu, Xiance Si, Jeremy Greer, Guolong Su, Martin Polacek, Raphaël~Lopez Kaufman, Simon Tokumine, Hexiang Hu, Elena Buchatskaya, Yingjie Miao, Mohamed Elhawaty, Aditya Siddhant, Nenad Tomasev, Jinwei Xing, Christina Greer, Helen Miller, Shereen Ashraf, Aurko Roy, Zizhao Zhang, Ada Ma, Angelos Filos, Milos Besta, Rory Blevins, Ted Klimenko, Chih-Kuan Yeh, Soravit Changpinyo, Jiaqi Mu, Oscar Chang, Mantas Pajarskas, Carrie Muir, Vered Cohen, Charline~Le Lan, Krishna Haridasan, Amit Marathe, Steven Hansen, Sholto Douglas, Rajkumar Samuel, Mingqiu Wang, Sophia Austin, Chang Lan, Jiepu Jiang, Justin Chiu, Jaime~Alonso Lorenzo, Lars~Lowe Sjösund, Sébastien Cevey, Zach Gleicher, Thi Avrahami, Anudhyan Boral, Hansa Srinivasan, Vittorio Selo, Rhys May, Konstantinos Aisopos, Léonard Hussenot, Livio~Baldini Soares, Kate Baumli, Michael~B. Chang, Adrià Recasens, Ben Caine, Alexander Pritzel, Filip Pavetic,
  Fabio Pardo, Anita Gergely, Justin Frye, Vinay Ramasesh, Dan Horgan, Kartikeya Badola, Nora Kassner, Subhrajit Roy, Ethan Dyer, Víctor~Campos Campos, Alex Tomala, Yunhao Tang, Dalia~El Badawy, Elspeth White, Basil Mustafa, Oran Lang, Abhishek Jindal, Sharad Vikram, Zhitao Gong, Sergi Caelles, Ross Hemsley, Gregory Thornton, Fangxiaoyu Feng, Wojciech Stokowiec, Ce~Zheng, Phoebe Thacker, Çağlar Ünlü, Zhishuai Zhang, Mohammad Saleh, James Svensson, Max Bileschi, Piyush Patil, Ankesh Anand, Roman Ring, Katerina Tsihlas, Arpi Vezer, Marco Selvi, Toby Shevlane, Mikel Rodriguez, Tom Kwiatkowski, Samira Daruki, Keran Rong, Allan Dafoe, Nicholas FitzGerald, Keren Gu-Lemberg, Mina Khan, Lisa~Anne Hendricks, Marie Pellat, Vladimir Feinberg, James Cobon-Kerr, Tara Sainath, Maribeth Rauh, Sayed~Hadi Hashemi, Richard Ives, Yana Hasson, Eric Noland, Yuan Cao, Nathan Byrd, Le~Hou, Qingze Wang, Thibault Sottiaux, Michela Paganini, Jean-Baptiste Lespiau, Alexandre Moufarek, Samer Hassan, Kaushik Shivakumar, Joost van
  Amersfoort, Amol Mandhane, Pratik Joshi, Anirudh Goyal, Matthew Tung, Andrew Brock, Hannah Sheahan, Vedant Misra, Cheng Li, Nemanja Rakićević, Mostafa Dehghani, Fangyu Liu, Sid Mittal, Junhyuk Oh, Seb Noury, Eren Sezener, Fantine Huot, Matthew Lamm, Nicola~De Cao, Charlie Chen, Sidharth Mudgal, Romina Stella, Kevin Brooks, Gautam Vasudevan, Chenxi Liu, Mainak Chain, Nivedita Melinkeri, Aaron Cohen, Venus Wang, Kristie Seymore, Sergey Zubkov, Rahul Goel, Summer Yue, Sai Krishnakumaran, Brian Albert, Nate Hurley, Motoki Sano, Anhad Mohananey, Jonah Joughin, Egor Filonov, Tomasz Kępa, Yomna Eldawy, Jiawern Lim, Rahul Rishi, Shirin Badiezadegan, Taylor Bos, Jerry Chang, Sanil Jain, Sri Gayatri~Sundara Padmanabhan, Subha Puttagunta, Kalpesh Krishna, Leslie Baker, Norbert Kalb, Vamsi Bedapudi, Adam Kurzrok, Shuntong Lei, Anthony Yu, Oren Litvin, Xiang Zhou, Zhichun Wu, Sam Sobell, Andrea Siciliano, Alan Papir, Robby Neale, Jonas Bragagnolo, Tej Toor, Tina Chen, Valentin Anklin, Feiran Wang, Richie Feng, Milad
  Gholami, Kevin Ling, Lijuan Liu, Jules Walter, Hamid Moghaddam, Arun Kishore, Jakub Adamek, Tyler Mercado, Jonathan Mallinson, Siddhinita Wandekar, Stephen Cagle, Eran Ofek, Guillermo Garrido, Clemens Lombriser, Maksim Mukha, Botu Sun, Hafeezul~Rahman Mohammad, Josip Matak, Yadi Qian, Vikas Peswani, Pawel Janus, Quan Yuan, Leif Schelin, Oana David, Ankur Garg, Yifan He, Oleksii Duzhyi, Anton Älgmyr, Timothée Lottaz, Qi~Li, Vikas Yadav, Luyao Xu, Alex Chinien, Rakesh Shivanna, Aleksandr Chuklin, Josie Li, Carrie Spadine, Travis Wolfe, Kareem Mohamed, Subhabrata Das, Zihang Dai, Kyle He, Daniel von Dincklage, Shyam Upadhyay, Akanksha Maurya, Luyan Chi, Sebastian Krause, Khalid Salama, Pam~G Rabinovitch, Pavan Kumar~Reddy M, Aarush Selvan, Mikhail Dektiarev, Golnaz Ghiasi, Erdem Guven, Himanshu Gupta, Boyi Liu, Deepak Sharma, Idan~Heimlich Shtacher, Shachi Paul, Oscar Akerlund, François-Xavier Aubet, Terry Huang, Chen Zhu, Eric Zhu, Elico Teixeira, Matthew Fritze, Francesco Bertolini, Liana-Eleonora
  Marinescu, Martin Bölle, Dominik Paulus, Khyatti Gupta, Tejasi Latkar, Max Chang, Jason Sanders, Roopa Wilson, Xuewei Wu, Yi-Xuan Tan, Lam~Nguyen Thiet, Tulsee Doshi, Sid Lall, Swaroop Mishra, Wanming Chen, Thang Luong, Seth Benjamin, Jasmine Lee, Ewa Andrejczuk, Dominik Rabiej, Vipul Ranjan, Krzysztof Styrc, Pengcheng Yin, Jon Simon, Malcolm~Rose Harriott, Mudit Bansal, Alexei Robsky, Geoff Bacon, David Greene, Daniil Mirylenka, Chen Zhou, Obaid Sarvana, Abhimanyu Goyal, Samuel Andermatt, Patrick Siegler, Ben Horn, Assaf Israel, Francesco Pongetti, Chih-Wei~"Louis" Chen, Marco Selvatici, Pedro Silva, Kathie Wang, Jackson Tolins, Kelvin Guu, Roey Yogev, Xiaochen Cai, Alessandro Agostini, Maulik Shah, Hung Nguyen, Noah~Ó Donnaile, Sébastien Pereira, Linda Friso, Adam Stambler, Adam Kurzrok, Chenkai Kuang, Yan Romanikhin, Mark Geller, ZJ~Yan, Kane Jang, Cheng-Chun Lee, Wojciech Fica, Eric Malmi, Qijun Tan, Dan Banica, Daniel Balle, Ryan Pham, Yanping Huang, Diana Avram, Hongzhi Shi, Jasjot Singh, Chris
  Hidey, Niharika Ahuja, Pranab Saxena, Dan Dooley, Srividya~Pranavi Potharaju, Eileen O'Neill, Anand Gokulchandran, Ryan Foley, Kai Zhao, Mike Dusenberry, Yuan Liu, Pulkit Mehta, Ragha Kotikalapudi, Chalence Safranek-Shrader, Andrew Goodman, Joshua Kessinger, Eran Globen, Prateek Kolhar, Chris Gorgolewski, Ali Ibrahim, Yang Song, Ali Eichenbaum, Thomas Brovelli, Sahitya Potluri, Preethi Lahoti, Cip Baetu, Ali Ghorbani, Charles Chen, Andy Crawford, Shalini Pal, Mukund Sridhar, Petru Gurita, Asier Mujika, Igor Petrovski, Pierre-Louis Cedoz, Chenmei Li, Shiyuan Chen, Niccolò~Dal Santo, Siddharth Goyal, Jitesh Punjabi, Karthik Kappaganthu, Chester Kwak, Pallavi LV, Sarmishta Velury, Himadri Choudhury, Jamie Hall, Premal Shah, Ricardo Figueira, Matt Thomas, Minjie Lu, Ting Zhou, Chintu Kumar, Thomas Jurdi, Sharat Chikkerur, Yenai Ma, Adams Yu, Soo Kwak, Victor Ähdel, Sujeevan Rajayogam, Travis Choma, Fei Liu, Aditya Barua, Colin Ji, Ji~Ho Park, Vincent Hellendoorn, Alex Bailey, Taylan Bilal, Huanjie Zhou,
  Mehrdad Khatir, Charles Sutton, Wojciech Rzadkowski, Fiona Macintosh, Konstantin Shagin, Paul Medina, Chen Liang, Jinjing Zhou, Pararth Shah, Yingying Bi, Attila Dankovics, Shipra Banga, Sabine Lehmann, Marissa Bredesen, Zifan Lin, John~Eric Hoffmann, Jonathan Lai, Raynald Chung, Kai Yang, Nihal Balani, Arthur Bražinskas, Andrei Sozanschi, Matthew Hayes, Héctor~Fernández Alcalde, Peter Makarov, Will Chen, Antonio Stella, Liselotte Snijders, Michael Mandl, Ante Kärrman, Paweł Nowak, Xinyi Wu, Alex Dyck, Krishnan Vaidyanathan, Raghavender R, Jessica Mallet, Mitch Rudominer, Eric Johnston, Sushil Mittal, Akhil Udathu, Janara Christensen, Vishal Verma, Zach Irving, Andreas Santucci, Gamaleldin Elsayed, Elnaz Davoodi, Marin Georgiev, Ian Tenney, Nan Hua, Geoffrey Cideron, Edouard Leurent, Mahmoud Alnahlawi, Ionut Georgescu, Nan Wei, Ivy Zheng, Dylan Scandinaro, Heinrich Jiang, Jasper Snoek, Mukund Sundararajan, Xuezhi Wang, Zack Ontiveros, Itay Karo, Jeremy Cole, Vinu Rajashekhar, Lara Tumeh, Eyal
  Ben-David, Rishub Jain, Jonathan Uesato, Romina Datta, Oskar Bunyan, Shimu Wu, John Zhang, Piotr Stanczyk, Ye~Zhang, David Steiner, Subhajit Naskar, Michael Azzam, Matthew Johnson, Adam Paszke, Chung-Cheng Chiu, Jaume~Sanchez Elias, Afroz Mohiuddin, Faizan Muhammad, Jin Miao, Andrew Lee, Nino Vieillard, Jane Park, Jiageng Zhang, Jeff Stanway, Drew Garmon, Abhijit Karmarkar, Zhe Dong, Jong Lee, Aviral Kumar, Luowei Zhou, Jonathan Evens, William Isaac, Geoffrey Irving, Edward Loper, Michael Fink, Isha Arkatkar, Nanxin Chen, Izhak Shafran, Ivan Petrychenko, Zhe Chen, Johnson Jia, Anselm Levskaya, Zhenkai Zhu, Peter Grabowski, Yu~Mao, Alberto Magni, Kaisheng Yao, Javier Snaider, Norman Casagrande, Evan Palmer, Paul Suganthan, Alfonso Castaño, Irene Giannoumis, Wooyeol Kim, Mikołaj Rybiński, Ashwin Sreevatsa, Jennifer Prendki, David Soergel, Adrian Goedeckemeyer, Willi Gierke, Mohsen Jafari, Meenu Gaba, Jeremy Wiesner, Diana~Gage Wright, Yawen Wei, Harsha Vashisht, Yana Kulizhskaya, Jay Hoover, Maigo Le,
  Lu~Li, Chimezie Iwuanyanwu, Lu~Liu, Kevin Ramirez, Andrey Khorlin, Albert Cui, Tian LIN, Marcus Wu, Ricardo Aguilar, Keith Pallo, Abhishek Chakladar, Ginger Perng, Elena~Allica Abellan, Mingyang Zhang, Ishita Dasgupta, Nate Kushman, Ivo Penchev, Alena Repina, Xihui Wu, Tom van~der Weide, Priya Ponnapalli, Caroline Kaplan, Jiri Simsa, Shuangfeng Li, Olivier Dousse, Fan Yang, Jeff Piper, Nathan Ie, Rama Pasumarthi, Nathan Lintz, Anitha Vijayakumar, Daniel Andor, Pedro Valenzuela, Minnie Lui, Cosmin Paduraru, Daiyi Peng, Katherine Lee, Shuyuan Zhang, Somer Greene, Duc~Dung Nguyen, Paula Kurylowicz, Cassidy Hardin, Lucas Dixon, Lili Janzer, Kiam Choo, Ziqiang Feng, Biao Zhang, Achintya Singhal, Dayou Du, Dan McKinnon, Natasha Antropova, Tolga Bolukbasi, Orgad Keller, David Reid, Daniel Finchelstein, Maria~Abi Raad, Remi Crocker, Peter Hawkins, Robert Dadashi, Colin Gaffney, Ken Franko, Anna Bulanova, Rémi Leblond, Shirley Chung, Harry Askham, Luis~C. Cobo, Kelvin Xu, Felix Fischer, Jun Xu, Christina Sorokin,
  Chris Alberti, Chu-Cheng Lin, Colin Evans, Alek Dimitriev, Hannah Forbes, Dylan Banarse, Zora Tung, Mark Omernick, Colton Bishop, Rachel Sterneck, Rohan Jain, Jiawei Xia, Ehsan Amid, Francesco Piccinno, Xingyu Wang, Praseem Banzal, Daniel~J. Mankowitz, Alex Polozov, Victoria Krakovna, Sasha Brown, MohammadHossein Bateni, Dennis Duan, Vlad Firoiu, Meghana Thotakuri, Tom Natan, Matthieu Geist, Ser tan Girgin, Hui Li, Jiayu Ye, Ofir Roval, Reiko Tojo, Michael Kwong, James Lee-Thorp, Christopher Yew, Danila Sinopalnikov, Sabela Ramos, John Mellor, Abhishek Sharma, Kathy Wu, David Miller, Nicolas Sonnerat, Denis Vnukov, Rory Greig, Jennifer Beattie, Emily Caveness, Libin Bai, Julian Eisenschlos, Alex Korchemniy, Tomy Tsai, Mimi Jasarevic, Weize Kong, Phuong Dao, Zeyu Zheng, Frederick Liu, Fan Yang, Rui Zhu, Tian~Huey Teh, Jason Sanmiya, Evgeny Gladchenko, Nejc Trdin, Daniel Toyama, Evan Rosen, Sasan Tavakkol, Linting Xue, Chen Elkind, Oliver Woodman, John Carpenter, George Papamakarios, Rupert Kemp, Sushant
  Kafle, Tanya Grunina, Rishika Sinha, Alice Talbert, Diane Wu, Denese Owusu-Afriyie, Cosmo Du, Chloe Thornton, Jordi Pont-Tuset, Pradyumna Narayana, Jing Li, Saaber Fatehi, John Wieting, Omar Ajmeri, Benigno Uria, Yeongil Ko, Laura Knight, Amélie Héliou, Ning Niu, Shane Gu, Chenxi Pang, Yeqing Li, Nir Levine, Ariel Stolovich, Rebeca Santamaria-Fernandez, Sonam Goenka, Wenny Yustalim, Robin Strudel, Ali Elqursh, Charlie Deck, Hyo Lee, Zonglin Li, Kyle Levin, Raphael Hoffmann, Dan Holtmann-Rice, Olivier Bachem, Sho Arora, Christy Koh, Soheil~Hassas Yeganeh, Siim Põder, Mukarram Tariq, Yanhua Sun, Lucian Ionita, Mojtaba Seyedhosseini, Pouya Tafti, Zhiyu Liu, Anmol Gulati, Jasmine Liu, Xinyu Ye, Bart Chrzaszcz, Lily Wang, Nikhil Sethi, Tianrun Li, Ben Brown, Shreya Singh, Wei Fan, Aaron Parisi, Joe Stanton, Vinod Koverkathu, Christopher~A. Choquette-Choo, Yunjie Li, TJ~Lu, Abe Ittycheriah, Prakash Shroff, Mani Varadarajan, Sanaz Bahargam, Rob Willoughby, David Gaddy, Guillaume Desjardins, Marco Cornero, Brona
  Robenek, Bhavishya Mittal, Ben Albrecht, Ashish Shenoy, Fedor Moiseev, Henrik Jacobsson, Alireza Ghaffarkhah, Morgane Rivière, Alanna Walton, Clément Crepy, Alicia Parrish, Zongwei Zhou, Clement Farabet, Carey Radebaugh, Praveen Srinivasan, Claudia van~der Salm, Andreas Fidjeland, Salvatore Scellato, Eri Latorre-Chimoto, Hanna Klimczak-Plucińska, David Bridson, Dario de~Cesare, Tom Hudson, Piermaria Mendolicchio, Lexi Walker, Alex Morris, Matthew Mauger, Alexey Guseynov, Alison Reid, Seth Odoom, Lucia Loher, Victor Cotruta, Madhavi Yenugula, Dominik Grewe, Anastasia Petrushkina, Tom Duerig, Antonio Sanchez, Steve Yadlowsky, Amy Shen, Amir Globerson, Lynette Webb, Sahil Dua, Dong Li, Surya Bhupatiraju, Dan Hurt, Haroon Qureshi, Ananth Agarwal, Tomer Shani, Matan Eyal, Anuj Khare, Shreyas~Rammohan Belle, Lei Wang, Chetan Tekur, Mihir~Sanjay Kale, Jinliang Wei, Ruoxin Sang, Brennan Saeta, Tyler Liechty, Yi~Sun, Yao Zhao, Stephan Lee, Pandu Nayak, Doug Fritz, Manish~Reddy Vuyyuru, John Aslanides, Nidhi Vyas,
  Martin Wicke, Xiao Ma, Evgenii Eltyshev, Nina Martin, Hardie Cate, James Manyika, Keyvan Amiri, Yelin Kim, Xi~Xiong, Kai Kang, Florian Luisier, Nilesh Tripuraneni, David Madras, Mandy Guo, Austin Waters, Oliver Wang, Joshua Ainslie, Jason Baldridge, Han Zhang, Garima Pruthi, Jakob Bauer, Feng Yang, Riham Mansour, Jason Gelman, Yang Xu, George Polovets, Ji~Liu, Honglong Cai, Warren Chen, XiangHai Sheng, Emily Xue, Sherjil Ozair, Christof Angermueller, Xiaowei Li, Anoop Sinha, Weiren Wang, Julia Wiesinger, Emmanouil Koukoumidis, Yuan Tian, Anand Iyer, Madhu Gurumurthy, Mark Goldenson, Parashar Shah, MK~Blake, Hongkun Yu, Anthony Urbanowicz, Jennimaria Palomaki, Chrisantha Fernando, Ken Durden, Harsh Mehta, Nikola Momchev, Elahe Rahimtoroghi, Maria Georgaki, Amit Raul, Sebastian Ruder, Morgan Redshaw, Jinhyuk Lee, Denny Zhou, Komal Jalan, Dinghua Li, Blake Hechtman, Parker Schuh, Milad Nasr, Kieran Milan, Vladimir Mikulik, Juliana Franco, Tim Green, Nam Nguyen, Joe Kelley, Aroma Mahendru, Andrea Hu, Joshua
  Howland, Ben Vargas, Jeffrey Hui, Kshitij Bansal, Vikram Rao, Rakesh Ghiya, Emma Wang, Ke~Ye, Jean~Michel Sarr, Melanie~Moranski Preston, Madeleine Elish, Steve Li, Aakash Kaku, Jigar Gupta, Ice Pasupat, Da-Cheng Juan, Milan Someswar, Tejvi M., Xinyun Chen, Aida Amini, Alex Fabrikant, Eric Chu, Xuanyi Dong, Amruta Muthal, Senaka Buthpitiya, Sarthak Jauhari, Nan Hua, Urvashi Khandelwal, Ayal Hitron, Jie Ren, Larissa Rinaldi, Shahar Drath, Avigail Dabush, Nan-Jiang Jiang, Harshal Godhia, Uli Sachs, Anthony Chen, Yicheng Fan, Hagai Taitelbaum, Hila Noga, Zhuyun Dai, James Wang, Chen Liang, Jenny Hamer, Chun-Sung Ferng, Chenel Elkind, Aviel Atias, Paulina Lee, Vít Listík, Mathias Carlen, Jan van~de Kerkhof, Marcin Pikus, Krunoslav Zaher, Paul Müller, Sasha Zykova, Richard Stefanec, Vitaly Gatsko, Christoph Hirnschall, Ashwin Sethi, Xingyu~Federico Xu, Chetan Ahuja, Beth Tsai, Anca Stefanoiu, Bo~Feng, Keshav Dhandhania, Manish Katyal, Akshay Gupta, Atharva Parulekar, Divya Pitta, Jing Zhao, Vivaan Bhatia,
  Yashodha Bhavnani, Omar Alhadlaq, Xiaolin Li, Peter Danenberg, Dennis Tu, Alex Pine, Vera Filippova, Abhipso Ghosh, Ben Limonchik, Bhargava Urala, Chaitanya~Krishna Lanka, Derik Clive, Yi~Sun, Edward Li, Hao Wu, Kevin Hongtongsak, Ianna Li, Kalind Thakkar, Kuanysh Omarov, Kushal Majmundar, Michael Alverson, Michael Kucharski, Mohak Patel, Mudit Jain, Maksim Zabelin, Paolo Pelagatti, Rohan Kohli, Saurabh Kumar, Joseph Kim, Swetha Sankar, Vineet Shah, Lakshmi Ramachandruni, Xiangkai Zeng, Ben Bariach, Laura Weidinger, Amar Subramanya, Sissie Hsiao, Demis Hassabis, Koray Kavukcuoglu, Adam Sadovsky, Quoc Le, Trevor Strohman, Yonghui Wu, Slav Petrov, Jeffrey Dean, and Oriol Vinyals.
\newblock Gemini: A family of highly capable multimodal models, 2024.

\bibitem[Tigges et~al.(2023)Tigges, Hollinsworth, Geiger, and Nanda]{tigges2023linear}
Curt Tigges, Oskar~John Hollinsworth, Atticus Geiger, and Neel Nanda.
\newblock Linear representations of sentiment in large language models, 2023.

\bibitem[Varma et~al.(2023)Varma, Shah, Kenton, Kramár, and Kumar]{varma2023explaining}
Vikrant Varma, Rohin Shah, Zachary Kenton, János Kramár, and Ramana Kumar.
\newblock Explaining grokking through circuit efficiency, 2023.

\bibitem[Vig et~al.(2020)Vig, Gehrmann, Belinkov, Qian, Nevo, Sakenis, Huang, Singer, and Shieber]{vig2020causal}
Jesse Vig, Sebastian Gehrmann, Yonatan Belinkov, Sharon Qian, Daniel Nevo, Simas Sakenis, Jason Huang, Yaron Singer, and Stuart Shieber.
\newblock Causal mediation analysis for interpreting neural nlp: The case of gender bias, 2020.

\bibitem[Voita et~al.(2023)Voita, Ferrando, and Nalmpantis]{voita2023neurons}
Elena Voita, Javier Ferrando, and Christoforos Nalmpantis.
\newblock Neurons in large language models: Dead, n-gram, positional, 2023.

\bibitem[Wang et~al.(2023)Wang, Variengien, Conmy, Shlegeris, and Steinhardt]{wang2023interpretability}
Kevin~Ro Wang, Alexandre Variengien, Arthur Conmy, Buck Shlegeris, and Jacob Steinhardt.
\newblock Interpretability in the wild: a circuit for indirect object identification in {GPT}-2 small.
\newblock In \emph{The Eleventh International Conference on Learning Representations}, 2023.
\newblock URL \url{https://openreview.net/forum?id=NpsVSN6o4ul}.

\bibitem[Warstadt et~al.(2020)Warstadt, Zhang, Li, Liu, and Bowman]{warstadt-etal-2020-learning}
Alex Warstadt, Yian Zhang, Xiaocheng Li, Haokun Liu, and Samuel~R. Bowman.
\newblock Learning which features matter: {R}o{BERT}a acquires a preference for linguistic generalizations (eventually).
\newblock In Bonnie Webber, Trevor Cohn, Yulan He, and Yang Liu, editors, \emph{Proceedings of the 2020 Conference on Empirical Methods in Natural Language Processing (EMNLP)}, pages 217--235, Online, November 2020. Association for Computational Linguistics.
\newblock \doi{10.18653/v1/2020.emnlp-main.16}.
\newblock URL \url{https://aclanthology.org/2020.emnlp-main.16}.

\bibitem[Welbl et~al.(2017)Welbl, Liu, and Gardner]{welbl-etal-2017-crowdsourcing}
Johannes Welbl, Nelson~F. Liu, and Matt Gardner.
\newblock Crowdsourcing multiple choice science questions.
\newblock In Leon Derczynski, Wei Xu, Alan Ritter, and Tim Baldwin, editors, \emph{Proceedings of the 3rd Workshop on Noisy User-generated Text}, pages 94--106, Copenhagen, Denmark, September 2017. Association for Computational Linguistics.
\newblock \doi{10.18653/v1/W17-4413}.
\newblock URL \url{https://aclanthology.org/W17-4413}.

\bibitem[Xia et~al.(2023)Xia, Artetxe, Zhou, Lin, Pasunuru, Chen, Zettlemoyer, and Stoyanov]{xia-etal-2023-training}
Mengzhou Xia, Mikel Artetxe, Chunting Zhou, Xi~Victoria Lin, Ramakanth Pasunuru, Danqi Chen, Luke Zettlemoyer, and Veselin Stoyanov.
\newblock Training trajectories of language models across scales.
\newblock In Anna Rogers, Jordan Boyd-Graber, and Naoaki Okazaki, editors, \emph{Proceedings of the 61st Annual Meeting of the Association for Computational Linguistics (Volume 1: Long Papers)}, pages 13711--13738, Toronto, Canada, July 2023. Association for Computational Linguistics.
\newblock \doi{10.18653/v1/2023.acl-long.767}.
\newblock URL \url{https://aclanthology.org/2023.acl-long.767}.

\bibitem[Zhang and Nanda(2024)]{zhang2024towards}
Fred Zhang and Neel Nanda.
\newblock Towards best practices of activation patching in language models: Metrics and methods.
\newblock In \emph{The Twelfth International Conference on Learning Representations}, 2024.
\newblock URL \url{https://openreview.net/forum?id=Hf17y6u9BC}.

\end{thebibliography}
\bibliographystyle{plainnat}
\newpage
\appendix
\section{Analysis of Task Circuits} \label{app:task-circuits}

\subsection{IOI Circuit \& Algorithmic Criteria}
To determine algorithmic consistency for the IOI circuit, we apply path patching as described in \Cref{app:manual-circuit-analysis} in addition to using the component scores described in \Cref{app:component-metrics}. These are used to set thresholds for classifying attention heads. Though component score thresholds can be arbitrary, applying them consistently across all model checkpoints allows us to see the degree of similarity involved with model behavior.

Concretely, we use the following metrics and thresholds:

\textbf{Direct-effect heads} We initially perform path-patching on all model attention heads, measuring their impact on the logit different after the final layer of the model. We then classify attention heads as name-mover heads (NMHs), negative name-mover heads, and copy suppression heads (CSHs) based on copy score (for NMHs) or CPSA (for CSHs) of $>10\%$, which yielded a small set of heads responsible for most of the direct effect. We measure the ratio of the absolute direct effect on logit difference for these heads vs. the total direct effect of all heads (including several unclassified heads) to obtain our first value.

Next, we conduct path-patching with NMHs as the receivers. This yields a set of heads that we then test for S2-inhibition (S2I) behavior, using \citeposs{wang2023interpretability} test for the effect of token signal vs. positional signal: does the ablation of these positional signal heads A). reduce the logit difference through the NMHs, B). reduce NMH attention (which determines what they copy) to the indirect object token, and C). increase attention to the subject tokens? If a head meets all of these conditions, we classify it as an S2I head, as it emits a signal used by the NMHs to decide what to copy. The total absolute effect of these heads on the NMHs is then divided by the total absolute effect of all heads on the NMHs, producing our second measurement.

Finally, we conduct path-patching with S2I heads as receivers. Here, we apply a simpler test since these heads can be quite diffuse throughout the model: Do the heads involved have above-average induction or duplicate-token scores? If so, we classify them as induction heads or duplicate token heads (confirming via manual examination of attention patterns and behavior), and divide the total absolute effect of these heads by the total absolute effect of all heads on the S2I heads, producing our third measurement.

These three metrics capture the extent to which known and classifiable model components contribute at each of the three primary levels of the IOI circuit. If the degree to which unknown or unclassified components contribute to any part of the circuit, we will see the corresponding score drop. As we see that in practice they tend to stay level, we conclude that there is a high degree of stability for this circuit.
\section{Other Circuit-Analysis Methods} \label{app:manual-circuit-analysis}
Circuit analysis can be conducted via a number of different methods; the method used to find the original IOI circuit (and that we use to verify algorithmic consistency in this task) is \citeposs{wang2023interpretability} \textbf{path-patching}. Path patching is a specialized form of activation patching, used to isolate and analyze the influence of individual model components on a given task. Starting with two datasets (identical except for the key detail we want to base our circuit on, such as the correct and incorrect names in the IOI task), \( x_{\text{orig}} \) and \( x_{\text{altered}} \), where \( x_{\text{altered}} \) is a counterfactual version of \( x_{\text{orig}} \), the technique involves a sender attention head \( h \) and a set of receiver nodes \( R \subseteq M \) within the model's computational graph \( M \). Initially, activations are recorded from both datasets. Subsequently, all attention heads except \( h \) are locked to their activations from \( x_{\text{orig}} \), while \( h \) is updated with its activation from \( x_{\text{altered}} \). This configuration allows for a forward pass on \( x_{\text{orig}} \), capturing intermediate activations for nodes \( r \in R \). A final forward pass on \( x_{\text{orig}} \) then patches \( R \) to these stored values, facilitating the assessment of \( h \)'s impact on the model's output.

Path patching aims to gauge the significance of the path \( h \rightarrow r \) by comparing the model's logit differences across multiple pairs \( (x_{\text{orig}}, x_{\text{new}}) \). By averaging these differences over many pairs, the method effectively measures the impact of specific paths on model performance, providing insights into the contributions of individual components to the overall task. The process is iterative, such that a practitioner would start by observing which nodes impact the logits directly, and then proceeding backwards to see what nodes affect those first direct-effect nodes, and so on.

\section{Additional Size, Similarity \& Change Rate Results} \label{app:change-graphs}

\begin{figure}
    \centering
       \includegraphics[width=\textwidth]{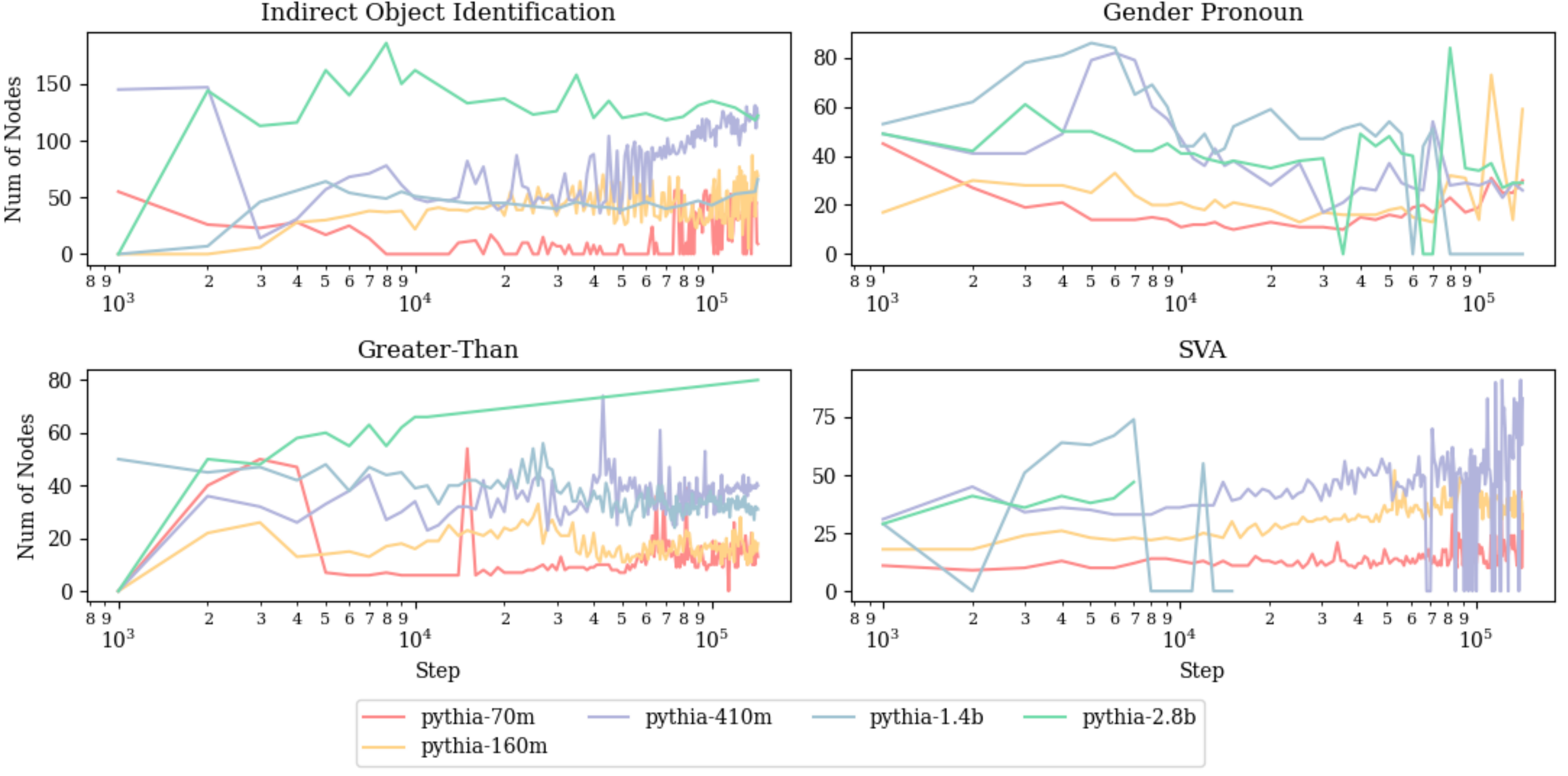}
    \caption{Number of Nodes in the circuits}
    \label{fig:circuit-size}
\end{figure}

\begin{figure}
    \centering
       \includegraphics[width=\textwidth]{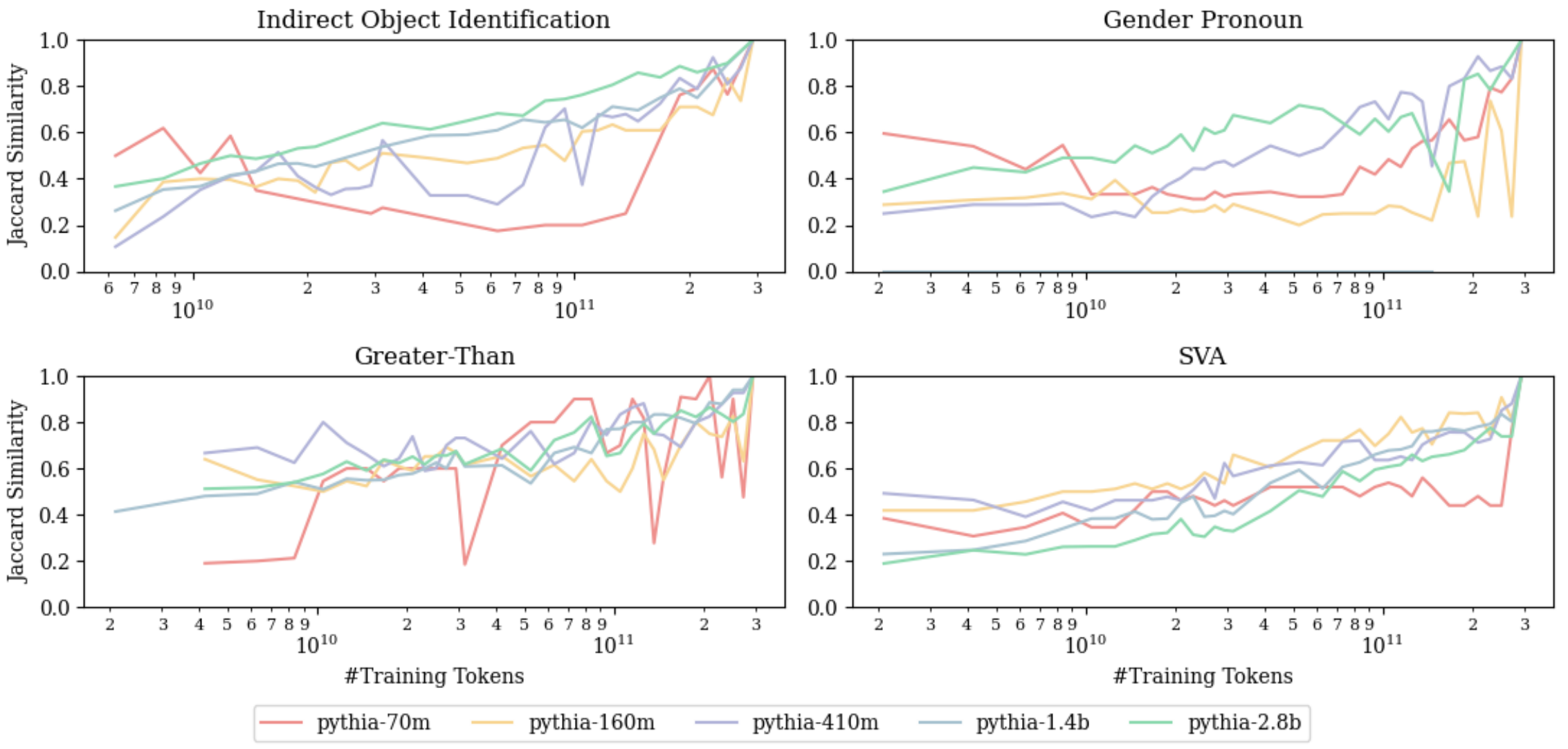}
    \caption{Node Jaccard Similarity of intermediate circuits with the circuit at the final checkpoint. Compared with the exponential weighted moving average (Figure 5, main text), we see more fluctuations, indicating that components swap during training at every checkpoint. Meanwhile, the upward trend indicates that circuits grow more and more similar to the final circuit during training.}
    \label{fig:njs}
\end{figure}

\begin{figure}
    \centering
       \includegraphics[width=\textwidth]{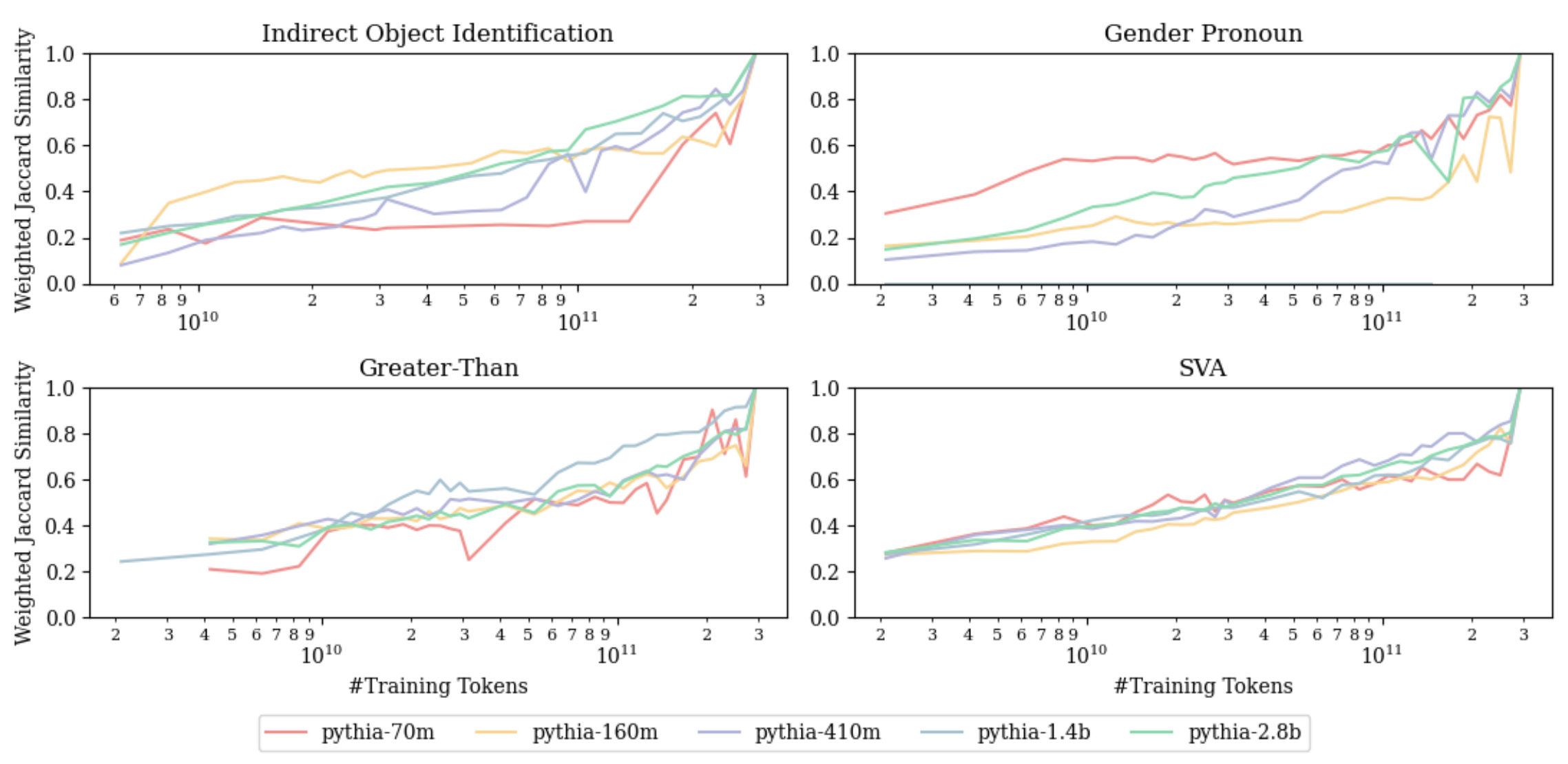}
    \caption{Edge Weighted Jaccard Similarity of the intermediate circuits with the circuit at the final checkpoint. The similarity slowly climbs and remains high at the end. Compared with the exact nodes, the weighted Jaccard Similarity of edges yields a smoother graph. It aligns with the conclusion that the model slowly drifts towards the final circuit along with components swapping during training}
    \label{fig:ewjs}
\end{figure}

We graph out the number of nodes needed to generate faithful circuits across different checkpoints on all of the four tasks. Here we can observe that the number of nodes is positively correlated to the sizes of the models. When the model size increases the model needs more heads of the same kind to complete the same tasks. In the case of IOI, we can see the pink line for pythia-70m is at the bottom with the least number of nodes and the green line of 2.8b is at the top with the most number of nodes. This signifies a diffusion of the roles in attention heads. Increases in model size do not necessarily help heads become more specialized in their roles; rather, in these circuits more heads will take on the same roles. 

\section{Component Metrics} \label{app:component-metrics}

In this paper, we follow the metrics from previous literature in \citet{wang2023interpretability} for name-mover heads, \citet{mcdougall2023copy} for copy suppression heads, \citep{olsson2022context} for induction heads, and \citep{gould2023successor} for successor heads.

\paragraph{Copy Score}
Following \citet{wang2023interpretability}, we check if the Name Mover Heads copy over the names across training time by using the same metrics- \textbf{copy score}. To validate the Name Mover Heads, we studied what values are written via the head's OV matrix. We take the state of the residual stream after the first layer of MLP on the specific name tokens. Then we multiply it with the OV matrix of the given heads, multiplied with the unembedding matrix and also the final layer norm. This simulates what will happen if the head attended perfectly to that token. We define copy score as the proportion of samples that contain the input name token in the top 5 logits. 

\paragraph{CSPA Score}
\citet{mcdougall2023copy} introduced a novel approach named copy suppression-preserving ablation (CSPA), designed to ablate all behaviors of a specified attention head except for those related to copy suppression. This method involves two distinct types of ablation: OV ablation and QK ablation. In the OV ablation process, the output of an attention head at a destination token \(D\) is represented as a weighted sum of result vectors from source tokens \(S\), with the weights corresponding to the attention probabilities from \(D\) to \(S\) \cite{elhage2021mathematical}. These vectors are then projected onto the unembedding vectors of their respective source tokens \(S\), retaining only their negative components. Meanwhile, QK ablation involves mean-ablating the result vectors from each source token \(S\), except for the top 5\% of source tokens that are most likely to be predicted at the destination token \(D\) based on the logit lens. For instance, in the phrase ``All’s fair in love and war,'' if the destination token \(D\) is ``and'' and the token ``love'' is a highly predicted follower of \(D\) and appears as a source token \(S\), the result vector from \(S\) is projected onto the unembedding vector for ``love,'' and everything else is mean-ablated. This demonstrates how the attention head in question suppresses the prediction of ``love.'' To evaluate the impact of the ablation, the token distribution output by the model for a given prompt (\(\pi\)) is compared with the distribution following an ablation (\(\pi_{Abl}\)) using KL divergence \(D_{KL}(\pi||\pi_{Abl})\). By averaging these values over the OpenWebText dataset, \(D_{CSPA}\) for CSPA and \(D_{MA}\) for a mean ablation baseline are obtained. The proportion of the effect explained is then calculated as \(1 - \frac{D_{CSPA}}{D_{MA}}\), with KL divergence chosen because a value of 0 indicates that the ablated and clean distributions are identical, implying that 100\% of the head’s effect is explained by the preserved components.

\paragraph{Previous Token Score}
The Previous Token Score measures how effectively each attention head attends to the immediately preceding token. To compute this, we use a diagonal extraction on the attention pattern matrices, offset by one position. This captures the attention weights directed to the token that precedes each token in the sequence. The scores are averaged over all batches and tokens, providing a mean score for each attention head across all layers.

\paragraph{Duplicate Token Score}
The Duplicate Token Score evaluates the propensity of each attention head to focus on duplicate tokens within a sequence. We achieve this by creating input sequences where the original tokens are repeated consecutively. The attention pattern matrices are then examined for their focus on tokens that are exactly a sequence length apart, indicating duplicate attention. The scores are calculated by averaging the attention weights along the specified diagonal, representing the attention paid to duplicate tokens.

\paragraph{Induction Head Score}
Based on the prefix matching score described by \citet{olsson2022context}, the Induction Head Score is designed to assess the ability of attention heads to engage in induction, where they predict the next token in a repeated sequence based on previously encountered patterns. To measure this, we generate sequences where a segment is repeated and compute the attention pattern matrices. We extract the diagonals offset by one less than the sequence length, capturing the attention from the end of the first segment to the start of the repeated segment. The mean attention scores along this diagonal provide the Induction Head Scores, averaged over all batches and tokens.

\paragraph{Succession Score} The succession score \citep{gould2023successor} measures the degree to which an attention head performs succession, upweighting ``2'' in response to ``1'', or ``May'' given the input ``April''. As \citeposs{gould2023successor} code is not publicly available, we re-implement their successor score as follows. We create a dataset of successor, consisting of numbers (in digit and written form), days of the week, and months. Then, we perform the following procedure from \citep{gould2023successor}. Letting $W_E$ and $W_U$ denote the embedding and unembedding matrices of the model under study, $MLP_0$ denote the first (zero-indexed) MLP layer, and $W_{OV}$ be the $OV$ matrix of the head under study. Then $M = W_UW_{OV} MLP_0(W_E)$ is a square matrix whose size is that of the model vocabulary; each row thereof indicates, for the corresponding word $x$ in the vocabulary, the degree to which an output word $y$ is upweighted by the head under study, when $x$ is in the input. For each $(x,y)$ pair in our dataset (e.g. (3,4) or (Tuesday, Wednesday)) we verify that $M[x][y] > M[x][y']$ for all $y'\neq y$ in our dataset; that is, we ensure that the correct answer is more highly upweighted than any of the other possible answers in our dataset. The succession score is the proportion of examples in which that is the case.

\section{Additional Evidence for Task-Dependent Learning Ceilings} \label{app:learning-ceiling-evidence}
\begin{figure}[htbp]
    \centering
    \subfigure[ARC Easy Accuracy]{
        \includegraphics[width=0.45\textwidth]{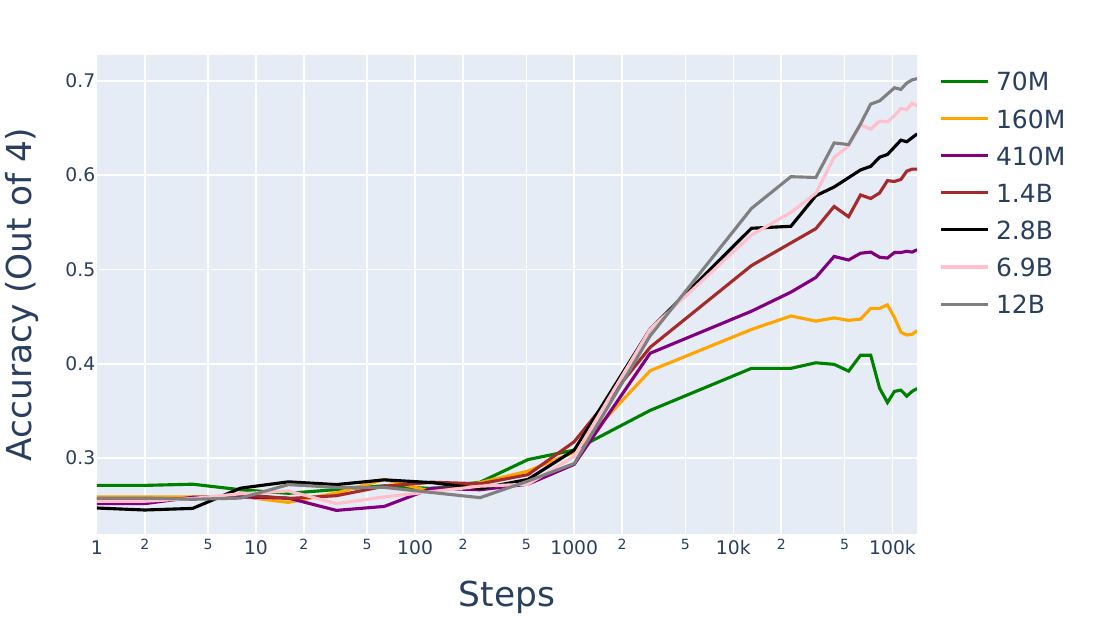}
    }
    \subfigure[SciQ Accuracy]{
        \includegraphics[width=0.45\textwidth]{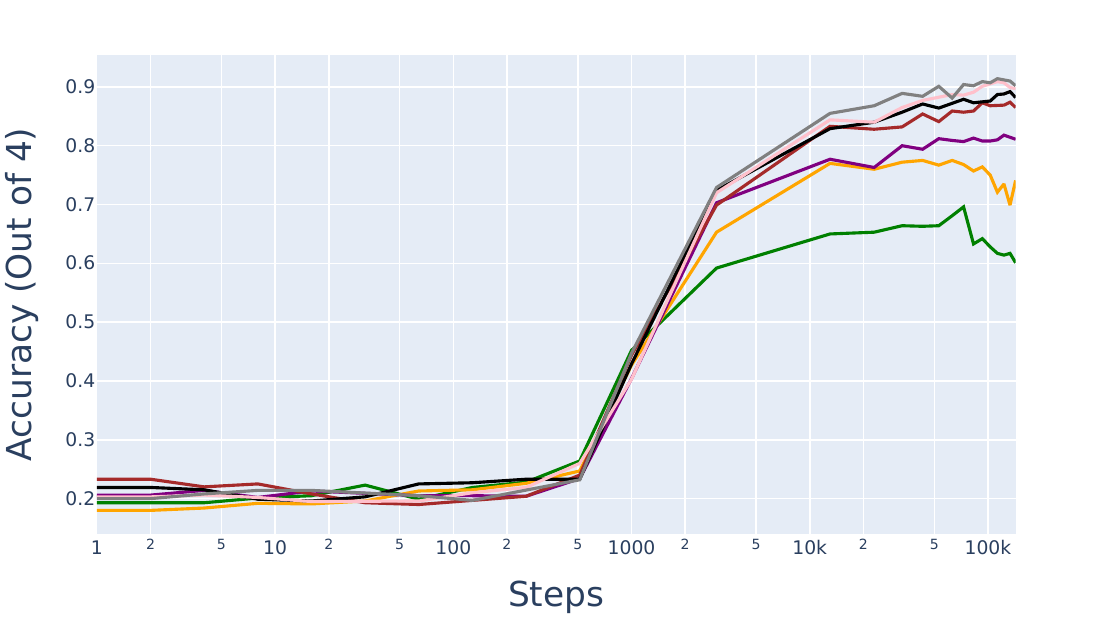}
    }
    \subfigure[PIQA Accuracy]{
        \includegraphics[width=0.45\textwidth]{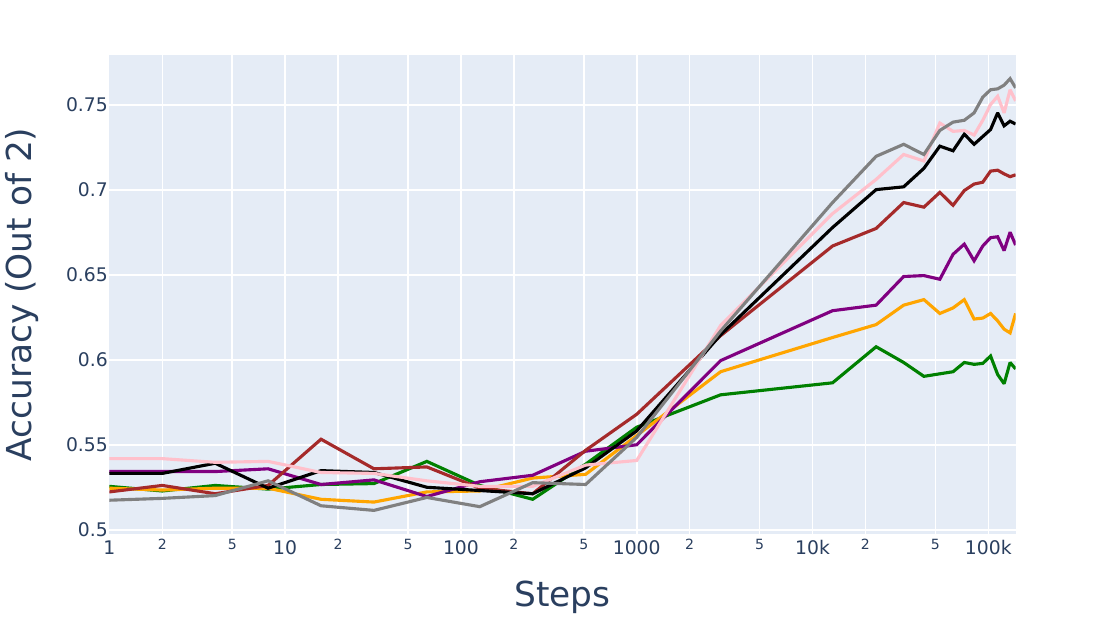}
    }
    \subfigure[Winogrande Accuracy]{
        \includegraphics[width=0.45\textwidth]{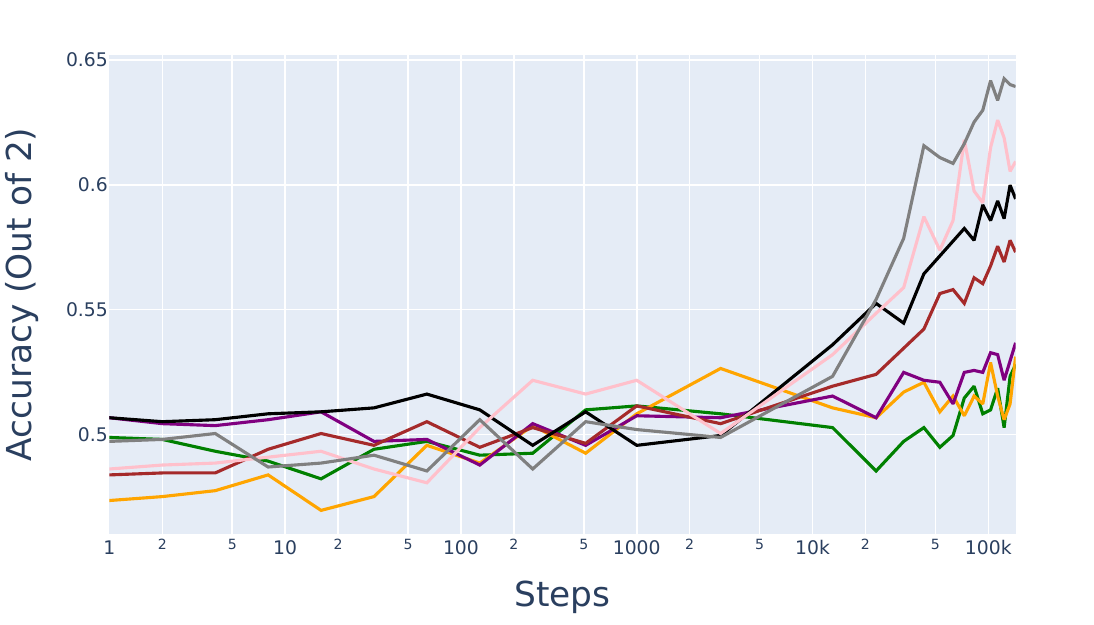}
    }
    \caption{Accuracy over training for four different datasets. Step numbers each represent approximately 2M tokens, so Step 1000 would be 2B tokens. We see that the rate of capability acquisition tends to approach an asymptote as models become larger.}
    \label{fig:pythia-evals}
\end{figure}

In addition to evaluations we performed ourselves, we also re-examined data collected during the Pythia training runs \citep{biderman2023pythia} on the SciQ \citep{welbl-etal-2017-crowdsourcing}, PIQA \citep{piqa}, WinoGrande \citep{winogrande}, and ARC Easy \citep{arc} datasets. Each of these consist of a wide range of questions with multiple-choice answers, and accuracy was evaluated on the basis of the top choice logit produced by the model. We find that performance acquisition rates on these tasks followed the same pattern we detected with our simpler task datasets--that is, task learning rate seemed to approach an asymptote as the models increased in size. We describe the datasets below and present the results in \Cref{fig:pythia-evals}.

\paragraph{SciQ}
        The Science Questions (SciQ) dataset \citep{welbl-etal-2017-crowdsourcing} consists of 13,679 crowdsourced multiple choice science exam questions ranging across physics, chemistry, biology, earth science, astronomy, and computer science. The questions cover a variety of complex reasoning skills such as causal reasoning, multi-hop inference, and understanding paragraph descriptions.

        \paragraph{PIQA}
        The Physical Interaction Question Answering (PIQA) dataset \citep{piqa} contains a total of 21k (across different subsets) multiple choice questions probing reasoning about basic physical commonsense knowledge. The questions test intuitive understanding of concepts like mass, volume, rigid objects, containment, stability, orientation, and more through grounded scenarios. Answering correctly requires applying physical reasoning.

        \paragraph{ARC Easy}
        The AI2 Reasoning Challenge (ARC) dataset \citep{arc} is a collection of 7,787 multiple choice science exam questions compiled from various grade-level sources, including a research partner of AI2. The questions cover diverse science topics and are structured as text-only prompts with 4 answer options. The ARC Easy subset consists of 5,197 of the relatively easier reasoning questions.

        \paragraph{Winogrande}
        The WinoGrande dataset \citep{winogrande} was inspired by the original Winograd Schema Challenge (WSC) and consists of 44k problems generated through crowdsourcing and systematic bias reduction algorithms. Most of these are relatively easy for humans, but often difficult for LLMs.

\section{Compute} \label{app:compute}
Experiments were conducted over two months a pod of 8 A40 GPUs, each with 50 GB of GPU RAM. As an upper bound, our experiments would require all of these GPUs to operate for a month to run all of our experiments, but in practice we did not require all GPUs running simultaneously. We estimate that 0.25 utilization of this pod would be required in practice to run these experiments.

\section{Licenses of Artifacts Used}\label{app:licenses}
The Pythia model suite is made available with an Apache 2.0 license. \citeposs{wang2023interpretability} IOI dataset and \citeposs{newman-etal-2021-refining} SVA dataset are released under an MIT license. The remaining datasets (Greater-Than and Gendered-Pronouns) are released without any license specified.
\end{document}